\documentclass[letterpaper]{article} 
\usepackage{aaai23}  
\usepackage{times}  
\usepackage{helvet}  
\usepackage{courier}  
\usepackage[hyphens]{url}  
\usepackage{graphicx} 
\usepackage{natbib}  
\usepackage{caption} 
\usepackage{algorithm}
\usepackage{algorithmic}
\usepackage{multirow}
\usepackage{color}
\usepackage{caption}
\usepackage{subfigure}
\usepackage{newfloat}
\usepackage{listings}

\definecolor{dkgreen}{RGB}{84,146,181}
\definecolor{gray}{rgb}{0.5,0.5,0.5}
\definecolor{mauve}{rgb}{0.58,0,0.82}

\usepackage[switch]{lineno}

\DeclareCaptionStyle{ruled}{labelfont=normalfont,labelsep=colon,strut=off} 

\DeclareCaptionFormat{mylst}{\hrule#1#2#3}
\usepackage{booktabs}

\title{High-Resolution GAN Inversion for Degraded Images in Large Diverse Datasets}
\author{
    Yanbo Wang\textsuperscript{\rm 1}\thanks{This work was done when Yanbo Wang was an intern at Tencent Youtu Lab.}, Chuming Lin\textsuperscript{\rm 2}, Donghao Luo\textsuperscript{\rm 2}, Ying Tai\textsuperscript{\rm 2}, Zhizhong Zhang\textsuperscript{\rm 1}\thanks{Corresponding author.}, Yuan Xie\textsuperscript{\rm 1}
}
\affiliations{
    \textsuperscript{\rm 1}School of Computer Science and Technology,
East China Normal University\\
    \textsuperscript{\rm 2}Tencent Youtu Lab\\

    51205901021@stu.ecnu.edu.cn, \{chuminglin, michaelluo, yingtai\}@tencent.com, \{zzzhang, yxie\}@cs.ecnu.edu.cn
}

\usepackage{bibentry}

\begin{document}

\maketitle

\begin{abstract}
The last decades are marked by massive and diverse image data, which shows increasingly high resolution and quality. However, some images we obtained may be corrupted, affecting the perception and the application of downstream tasks. A generic method for generating a high-quality image from the degraded one is in demand. In this paper, we present a novel GAN inversion framework that utilizes the powerful generative ability of StyleGAN-XL for this problem. To ease the inversion challenge with StyleGAN-XL, Clustering \& Regularize Inversion (CRI) is proposed. Specifically, the latent space is firstly divided into finer-grained sub-spaces by clustering. Instead of initializing the inversion with the average latent vector, we approximate a centroid latent vector from the clusters, which generates an image close to the input image. Then, an offset with a regularization term is introduced to keep the inverted latent vector within a certain range. We validate our CRI scheme on multiple restoration tasks (\textit{i.e.}, inpainting, colorization, and super-resolution) of complex natural images, and show preferable quantitative and qualitative results. We further demonstrate our technique is robust in terms of data and different GAN models. To our best knowledge, we are the first to adopt StyleGAN-XL for generating high-quality natural images from diverse degraded inputs. \emph{Code is available at \textcolor{magenta}{https://github.com/Booooooooooo/CRI}.}
\end{abstract}

\begin{figure*}
    \centering
    \includegraphics[width=0.95\linewidth]{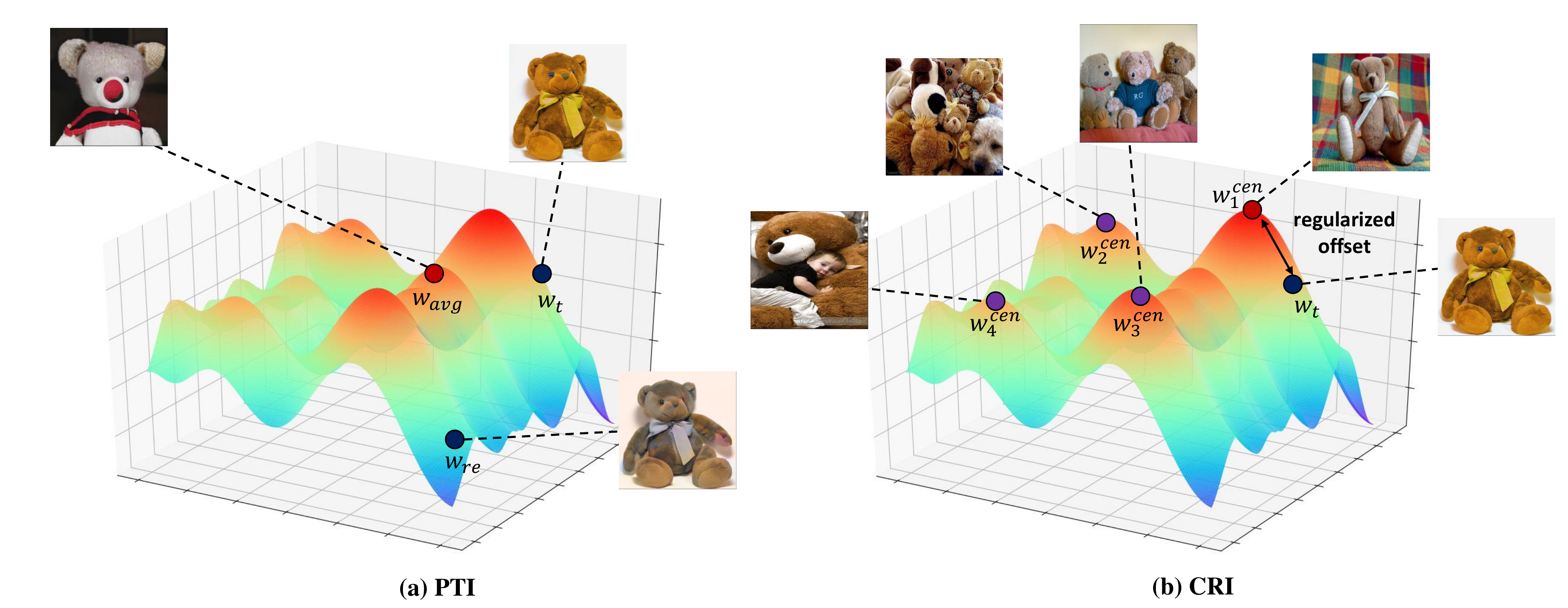}
    \caption{Comparison of existing GAN inversion methods and our proposed CRI in the latent space (\textit{e.g.}, $\mathcal{W}$ and $\mathcal{W}+$ space). (a) Existing inversion methods (\textit{e.g.}, PTI) begin the optimization from an average latent vector $w_{avg}$ of the class, and minimize the reconstruction loss. Due to the ill-posedness of image restoration, the resulted image $\phi_{Synthesis}(w_{re})$ is similar to the target image $\phi_{Synthesis}(w_t)$ but is visually unpleasant as it falls into the distribution margin. (b) Our CRI starts the optimization by finding the ``nearest'' centroid as the starting latent vector. $w^{cen}_1, w^{cen}_2, w^{cen}_3, w^{cen}_4$ are candidate centroid and $w^{cen}_1$ is the selected centroid. By bounding the scope of latent vectors with the regularized offset, the perceptual quality is guaranteed.} 
    \label{fig:method}
\end{figure*}

\section{Introduction}
With the mushroom development of imaging devices, images are now characterized by increasingly high resolution and diverse scenes. However, the quality of images on the Internet is uneven. For example, we may get images with low-resolution, images with missing parts, or gray-scale images. Therefore we hope to generate high-quality images directly from the degraded raw natural images. 

One feasible approach is to leverage the generative ability of generative adversarial networks (GAN) and conduct GAN inversion. The main idea is to ``invert'' the corrupted image back to the latent space of the pre-trained GAN and then reconstruct a restored image from the optimal latent~\cite{xia2022gan}. For a given degraded function $D(\cdot)$ and degraded image $I_d$, the latent vector $w_{re}$ is iterated through back-propagation by minimizing the reconstruction loss of $I_d$ and the degraded synthesis image $D(\phi_{Synthesis}(w_{re}))$, where $\phi_{Synthesis}$ is the synthesis layers of GAN. For example, PTI~\cite{roich2021pivotal} inverts high-quality face images to the latent space and uses the result for image attributes manipulation. Nevertheless, PTI cannot be generalized to degraded input images from natural scenes. On the other hand, DGP~\cite{pan2021exploiting}  provides compelling results to generate missing semantics, \textit{e.g.}, color, patch, resolution, for various images. But the scheme is limited to vanilla GAN models (\textit{i.e.}, BigGAN~\cite{brock2018large}) which only generate results with relatively low resolution (\textit{e.g.}, $128^2, 256^2$).

In this paper, we present an effective framework to generate high-resolution natural images from degraded inputs. We utilize the powerful capability of StyleGAN-XL~\cite{sauer2022stylegan}, the state-of-the-art model for large-scale image synthesis, to generate high-quality images from degraded images. However, simply applying existing inversion methods~\cite{pan2021exploiting, roich2021pivotal, menon2020pulse} on StyleGAN-XL pre-trained on ImageNet~\cite{krizhevsky2012imagenet} obtains unsatisfactory results and faces two challenges: 

(1) The latent space of StyleGAN-XL pre-trained on ImageNet is much larger and more complex compared to other popular GAN models like BigGAN and StyleGAN~\cite{karras2019stylev1}. Due to the diversity of data, the scale of the generator increases. For instance, for face images (\textit{i.e.}, FFHQ~\cite{karras2019stylev1}) StyleGAN2~\cite{karras2020stylev2} at resolution $1024^2$ only uses 18 latent vectors, while StyleGAN-XL trained on ImageNet at resolution $512^2$ needs 37 latent vectors since the data domain is more complicated. In the meantime, StyleGAN-XL employs an intermediate latent space compared to BigGAN which is the essence of its flexible editability but also adds to the inversion difficulty. \textit{For such a complex latent space, finding a good initialization of latent vector is quite important.} 

(2) The visual quality of the result is not explicitly imposed by the existing inversion methods. For example, images $I_{re}$ and $I_t$ generated by $w_{re}$ and $w_t$ in Figure \ref{fig:method}(a) result in the same degraded images ($D(I_{re})$ and $D(I_t)$). Minimizing the traditional pixel or feature loss as existing methods~\cite{roich2021pivotal} can achieve low distortion (indicated by metrics such as LPIPS~\cite{zhang2018unreasonable}) but neglect the perception (indicated by metrics such as NIQE~\cite{mittal2012making} and FID~\cite{heusel2017gans}). Consequently, some latent vectors may fall into the distribution margins of the $\mathcal{W}$ latent space of StyleGAN-XL (\textit{i.e.}, blue regions in Figure \ref{fig:method}), bringing poor and unpleasant results. \textit{Hence, a constraint specially designed from the perception perspective is in need.}

To tackle the aforementioned challenges, we proposed Clustering \& Regularize Inversion (CRI). We show diverse datasets such as ImageNet exhibit multi-modal characteristics, as shown in Figure \ref{fig:method}(b). For instance, images of the teddy bear class in ImageNet may contain one to many bears (\textit{i.e.}, $w^{cen}_1, w^{cen}_2, w^{cen}_3$) or even a person holding a teddy bear (\textit{i.e.}, $w^{cen}_4$). Therefore, instead of starting the optimization from an average latent vector like most existing methods, we choose a ``nearest'' centroid by clustering. The large and complex latent space is first clustered into finer-grained sub-spaces. The latent vector $w_k^{cen}$ (\textit{i.e.}, centroid) which generates the ``nearest'' image is selected as the starting point for inversion.  Meanwhile, inspired by the inherent distortion-perception trade-off in the latent space of StyleGAN~\cite{tov2021designing}, we improve the perceptual quality of results by constraining the scope of latent vectors. Specifically, an offset term $w^{off}$ with a regularization term is introduced. During the optimization, we keep the $w_k^{cen}$ frozen and iterate $w^{off}$ instead of optimizing the whole latent vector $w_k^{cen}+w^{off}$. By doing so, we explicitly bound the latent vectors to the high perception areas in the latent space. In short, our main contributions can be summarized as:
\begin{itemize}
    \item We fully explore the potential of GAN inversion for degraded natural images in diverse scenes. To our best knowledge, we are the first to adopt StyleGAN-XL to generate high-quality images from degraded inputs. 
    \item We propose a simple yet effective GAN inversion method to address the challenge of inversion with StyleGAN-XL. The key idea is to find a better starting latent vector and introduce a constraint on the image quality for optimization, summarized as Clustering \& Regularize.
    \item We conduct extensive experiments on restoration tasks with different datasets and GAN models. The experimental results demonstrate the superiority of our method against other inversion methods.
\end{itemize}

\section{Related Works}
\subsection{Style-based Generator}
The Style-based generator is first proposed by~\cite{karras2019stylev1}. A mapping module is introduced to map the random values (denoted by $z\in\mathcal{Z}$) to an intermediate latent space named $\mathcal{W}$ space. By feeding latent vectors $w\in\mathcal{W}$ to each synthesis layer, StyleGAN can control the attributes of the generated image. It quickly surpassed previous image generative models~\cite{brock2018large, karras2017progressive}, showing superior perceptual quality and variety. StyleGAN2~\cite{karras2020stylev2} introduced path length regularization and weight demodulation, which further improves the image quality. Recently, StyleGAN3~\cite{karras2021stylev3} addressed the aliasing problem and proposed a new architecture that boosts the generator to be fully equivariant to translation and rotation. 

\subsection{GAN Inversion}
GAN inversion aims at mapping a given image back into the latent space of a pre-trained GAN model, which can be viewed as the inverse problem of image synthesis. It was first introduced by~\cite{zhu2016generative}. Generally, inversion methods can be divided into optimization-based~\cite{creswell2018inverting, abdal2020image2stylegan++, roich2021pivotal} and learning-based~\cite{richardson2021encoding, dinh2022hyperinverter, tov2021designing, dong2021}. The optimization-based approaches directly update the latent vector by minimizing the reconstruction loss. Learning-based methods employ an encoder to learn the mapping from the given image to its corresponding latent. Recently, \cite{dong2021} studies the invertibility of an arbitrary pre-trained DNN using learning-based inversion. However, despite the gain in inference speed, the reconstruction quality of learning-based methods is often worse than that of optimization-based approaches. Also, the encoder cannot be adapted to out-of-domain data. Our proposed CRI is based on optimization as we find that the relatively easy optimization-based approaches still do not work well for complex scenes. 

Due to the many desirable properties of StyleGANs, recent works focus on conducting inversion with StyleGANs.
For StyleGANs, inversion is usually  conducted in the $\mathcal{W}$ space. It has been shown that the extended $\mathcal{W}+$ where different latent vectors are fed into each of the generator's layers is much more expressive and enables better image preservation~\cite{abdal2019image2stylegan}. As shown in \cite{tov2021designing}, there exist trade-offs between distortion, perceptual quality, and editability within the $\mathcal{W}+$ space. Recently, PTI~\cite{roich2021pivotal} proposed pivotal tuning to mitigate the distortion-editability trade-off for out-of-distribution images. 
In comparison, we consider GAN inversion as a tool for exploiting the GAN priors for restoration tasks. Finding a ``sweet spot'' in the distortion-perception trade-off rather than the distortion-editability trade-off is the primary purpose and challenge of our work. 
\subsection{Deep Priors in Image Restoration}
Image restoration is the task of recovering a clean image given its degraded version. Due to the powerful ability to generate photo-realistic images, many works exploit pre-trained GANs as priors to improve the quality~\cite{Kim2022BigColor, wei2022e2style, bartz+bethge2020model}.
Leveraging a pre-trained StyleGAN is very common in face restoration~\cite{menon2020pulse, yang2021gan}. 
As one representative work, PULSE~\cite{menon2020pulse} propose to solve face image super-resolution task with inversion on StyleGAN2. 
As for more complex real-world images, attempts are still limited to vanilla GANs (\textit{e.g.}, BigGAN).
DGP~\cite{pan2021exploiting} presents a relaxed formulation for mining the priors in BigGAN. The generator is jointly finetuned with the latent vectors. \cite{wu2021towards} `retrieves' color information encapsulated in BigGAN via an encoder and then incorporates these features with feature modulations. In this paper, we explore the potential of StyleGAN-XL for degraded images from natural complex scenes, achieving better results than existing GAN inversion methods. 

\section{Method}
Suppose the given input image $I_d$ is obtained via $I_d=D(I)$ where $I$ is the original high-quality image, $D(\cdot)$ is the degradation function. Our aim is to generate high-quality images. By utilizing StyleGAN-XL, the problem can be translated as finding a latent vector $w$ that satisfies:
\begin{equation}
    \label{eq:problem}
    w= \mathop{\arg\min}_wL(D(\phi_{Synthesis}(w)), I_d),
\end{equation}
where $\phi_{Synthesis}(\cdot)$ is the synthesis network of the pre-trained generator, $L(\cdot)$ is a distance metric in the image or feature space.

Although various attempt has been made~\cite{menon2020pulse, pan2021exploiting} to seek generative priors, generating high-resolution natural images remains challenging. Since the latent space is much more complex for StyleGAN-XL pre-trained on ImageNet, it is essential to provide additional help for the optimization process. We proposed a Clustering \& Regularize strategy to break through this problem. Clustering offers a better initial starting point for the optimization, while Regularize constrains the scope of $w$ during optimization. In this section, we will first give details about this strategy and then describe the overall pipeline. 


\subsection{Start from a Centroid}
One main challenge for restoring natural images with StyleGAN-XL lies in the complexity of the latent space. The latent vectors for StyleGAN-ImageNet present a multi-modal characteristic. Simply starting from an average latent vector becomes sub-optimal. Thus, we begin the optimization by finding a proper centroid as shown in Figure \ref{fig:method}(b). 

For a given class, we first randomly sample $M$ latent vectors $\{w_j\}^M_{j=1}$ using the pre-trained mapping module $\phi_{Mapping}$ of StyleGAN-XL:
\begin{equation}
\label{eq:map}
    \{w_j\}^M_{j=1} = \phi_{Mapping}(\{z_j\}^M_{j=1}, c),
\end{equation}
where $\{z_j\}^M_{j=1} \sim \mathcal{N}(0, 1)$ and $c$ is a one-hot vector of the given class.
$\{w_j\}^M_{j=1}$ are clustered into $N$ clusters, obtaining $N$ centroids $\{w_i^{cen}\}^N_{i=1}$ corresponding to $N$ center images:
\begin{equation}
    \{I^{cen}_i\}^N_{i=1} = \phi_{Synthesis}(\{w_i^{cen}\}^N_{i=1}). 
\end{equation}

Given degraded input image $I_d$, we measure the distance between $I_d$ and $\{I^{cen}_i\}^N_{i=1}$ in the feature space of a pre-trained model (e.g. VGG~\cite{simonyan2014very}). The corresponding $w^{cen}_k$ of the ``nearest'' center image is selected as the starting point for the subsequent optimization. 

\subsection{Regularized Offset}
Due to the ill-posedness of image restoration, we may find multiple latent vectors that satisfy Eq. \ref{eq:problem}. However, some of the generated images may be beyond the manifold of natural images. Applying the traditional loss on $I_d$ and $D(\phi_{synthesis}(w))$ does not explicitly guarantee the perception for inversion. We propose to improve the perceptual quality of the result images by constraining the scope of the latent vectors during optimization. 
Instead of optimizing $w$ directly, we introduce an offset term $w^{off}$ and keep the centroid $w_k^{cen}$ frozen. The output image is defined as:
\begin{equation}
    I_{syn} = \phi_{synthesis}(w^{cen}_k+w^{off}).
\end{equation}
Then we add a regularization term for $w^{off}$ 
\begin{equation}
    reg=||w^{off}||_2.
\end{equation}
This regularization term limits the range of the output latent $w=w^{cen}_k+w^{off}$ to a certain extent so that the output image does not exceed the natural manifold. By doing so, the latent is allowed to be gradually rectified towards the target while remaining high perceptual quality. 

\subsection{Loss Function}
The overall pipeline of CRI can be divided into two stages: the optimization stage and the finetune stage. As noted in DGP~\cite{pan2021exploiting}, a fixed generator is perhaps a crucial limitation for faithfully reconstructing unseen and complex images. However, we found that finetuning the generator together with the latent vector at the same time like DGP is sub-optimal for more sophisticated GANs. For StyleGANs with disentangled latent vectors, dividing optimization and finetune into two stages helps reduce the inversion difficulty.  

In the first stage, only the $w^{off}$ is optimized and the weights of the generator are frozen. 
The overall loss for the optimization stage can be defined as:
\begin{equation}
\resizebox{0.95\hsize}{!}{%
        $L_{op} = L_{LPIPS}(I_d, D(I_{syn})) + \lambda_1L_2(I_d, D(I_{syn})) + \lambda_2||w^{off}||_2,$%
}
\end{equation}
where $D(\cdot)$ is the degradation function, $\lambda_1$ and $\lambda_2$ are hyperparameters. As the noise input is removed in StyleGAN3, we do not optimize this term. However, the noise vector can be considered to improve the visual details. 

We use the pivotal tuning technique~\cite{roich2021pivotal} in the second stage, where the pivotal latent $w=w^{cen}_c+w^{off}$ is frozen. 
The generator is finetuned with the following loss:
\begin{equation}
\resizebox{0.95\hsize}{!}{%
        $L_{ft} = L_{LPIPS}(I_d, D(I_{syn})) + \lambda_{L2}L_2(I_d, D(I_{syn})) + \lambda_RL_R,$%
        }
\end{equation}
where $\lambda_{L2}$ and $\lambda_R$ are hyperparameters, $L_R$ is the locality regularization term defined as:
\begin{equation}
    L_R = L_{LPIPS}(x_r, x_r^*)+\lambda_{L2}^RL_{2}(x_r, x_r^*),
\end{equation} 
where $\lambda_{L2}^R$ is a hyperparameter, $x_r=\phi_{synthesis}(w_r; \theta)$ is generated with the original weights of the generator and $x_r^*=\phi_{synthesis}(w_r; \theta^*)$ is generated using the currently tuned ones, $w_r$ is the interpolated code between a random latent vector and the pivotal latent vector.
Pseudocode of CRI is summarized in Algorithm \ref{lst:listing}.

\begin{lstlisting}[language=python,caption={Pseudocode of CRI}, label=lst:listing, float=t]
# G: pre-trained GAN
# Id: degraded input image
# ci: the class index of the input image
# loss_fn_w, loss_fn_g: Equation 6, 7

# Clustering centroids
z_samples = randn(M, G.z_dim)
c_samples = one_hot(M, ci, G.c_dim)
w_samples = G.mapping(z_samples, c_samples)
km = KMeans(cluster_n).fit(w_samples)
centroids = km.cluster_centers_

# Estimating a centroid
center_images = G.synthesis(centroids)
center_features = VGG(center_images)
target_features = VGG(Id)
dis = L2(target_features, center_features)
w_st = centroids[dis.argmin(0)]

# Optimizing regularized offset
w_offset = zeros(w_st.shape)
w_offset.requires_grad_(True)
w_st.requires_grad_(False)
image = G.synthesis(w_st+w_offset)
loss = loss_fn_w(degrade(image), Id)
loss.backward()

# Finetune Generator
w_offset.requires_grad_(False)
G.requires_grad_(True)
image = G.synthesis(w_st+w_offset)
loss = loss_fn_g(degrade(image), Id)
loss.backward()

\end{lstlisting}
\section{Experiments}
In this section, we present comprehensive experiments to evaluate our method. We experiment with image inpainting, image colorization and super-resolution. We start by comparing with current inversion methods on StyleGAN-XL~\cite{sauer2022stylegan} pre-trained on ImageNet~\cite{krizhevsky2012imagenet}. We also compare the generating results of our method on StyleGAN-XL and DGP on BigGAN. For face images, we conduct experiments with StyleGAN2~\cite{karras2020stylev2} pre-trained on FFHQ~\cite{karras2019stylev1}, which shows the generalizability of our method. Then, we extend our method to out-of-domain images. Finally, we conduct an ablation study to justify the effectiveness of our inversion method. All the methods are inverted to $\mathcal{W} +$ space except PTI-colorization as we found $\mathcal{W}$ space generates better results. 

\subsection{Restoration on ImageNet}
We first compare our method with other GAN inversion methods~\cite{menon2020pulse, pan2021exploiting, roich2021pivotal} on ImageNet. We use StyleGAN-XL~\cite{sauer2022stylegan} as it is state-of-the-art on large-scale image synthesis. The resolution of the generated images is $512^2$. We invert 1000 images from the validation set of ImageNet, each from different classes, to quantitatively evaluate the methods.
The scale factor of the SR task is set to 4. 

\begin{figure*}[t]
    \centering
    \begin{minipage}{0.16\linewidth}
        \begin{center}
            Input
        \end{center}
        \includegraphics[width=1\linewidth]{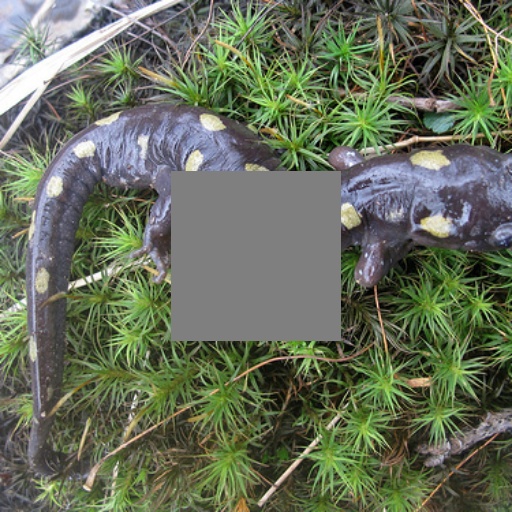}
        \includegraphics[width=1\linewidth]{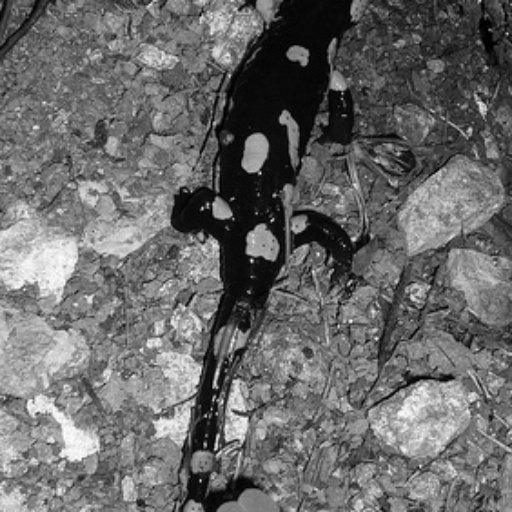}        \includegraphics[width=1\linewidth]{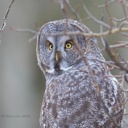}
    \end{minipage}
    \begin{minipage}{0.16\linewidth}
        \begin{center}
            PULSE
        \end{center}
        \includegraphics[width=1\linewidth]{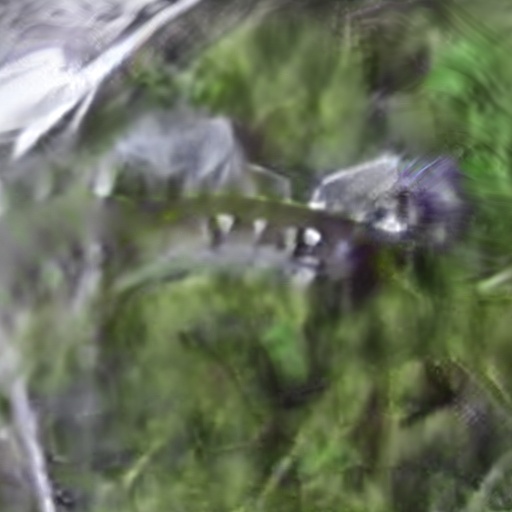}
        \includegraphics[width=1\linewidth]{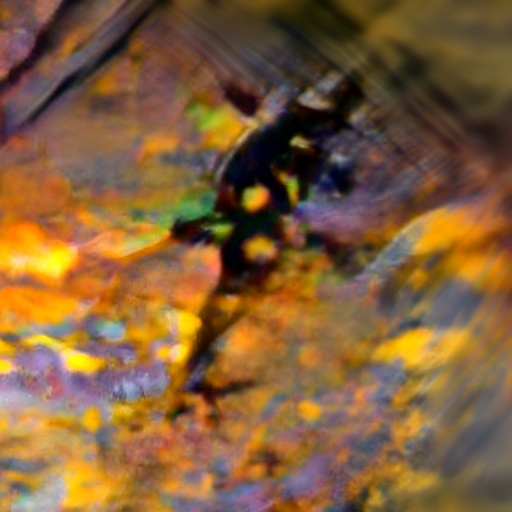}
        \includegraphics[width=1\linewidth]{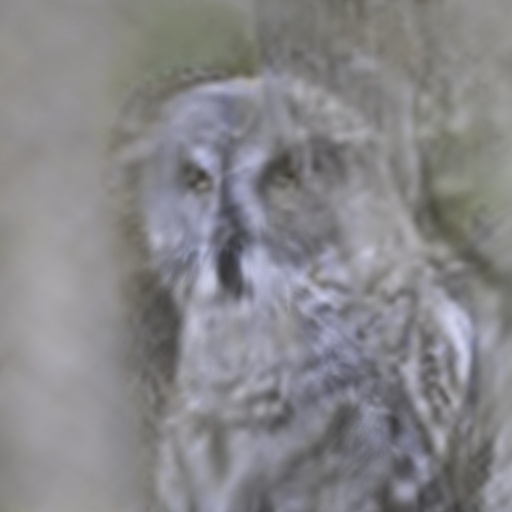}
    \end{minipage}
    \begin{minipage}{0.16\linewidth}
        \begin{center}
            DGP
        \end{center}
        \includegraphics[width=1\linewidth]{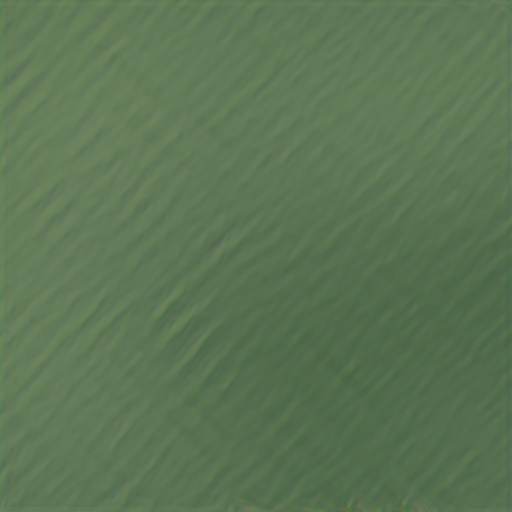}
        \includegraphics[width=1\linewidth]{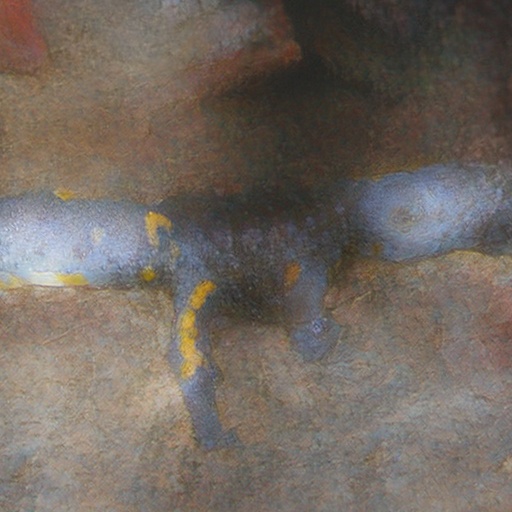}
        \includegraphics[width=1\linewidth]{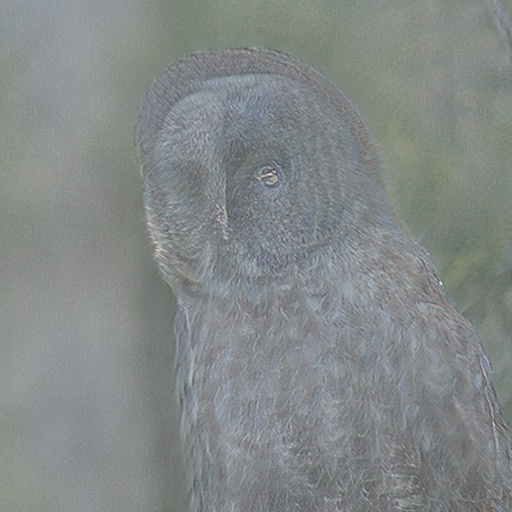}
    \end{minipage}
    \begin{minipage}{0.16\linewidth}
        \begin{center}
            PTI
        \end{center}
        \includegraphics[width=1\linewidth]{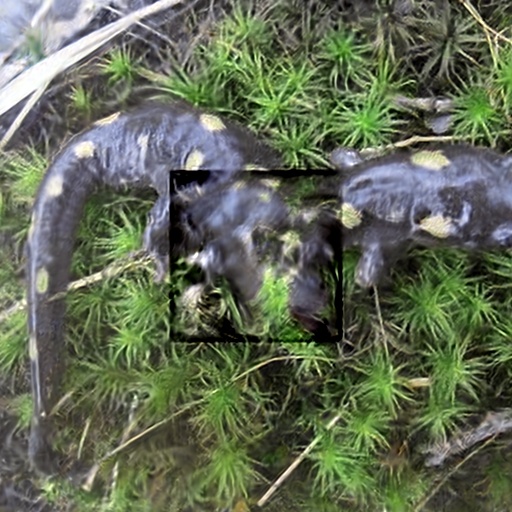}
        \includegraphics[width=1\linewidth]{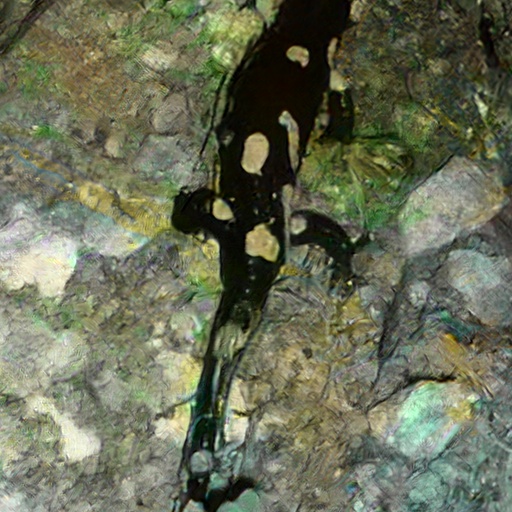}
        \includegraphics[width=1\linewidth]{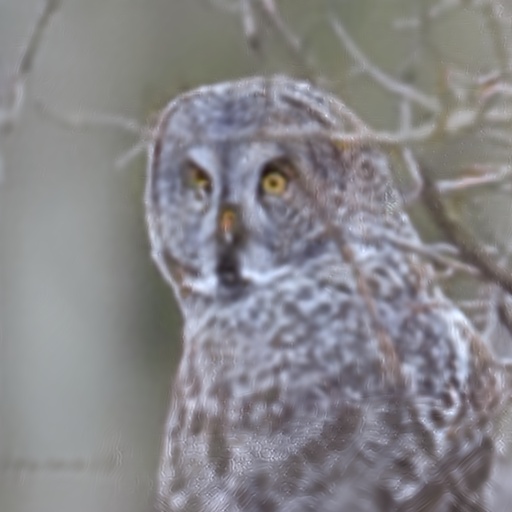}
    \end{minipage}
    \begin{minipage}{0.16\linewidth}
        \begin{center}
            Ours
        \end{center}
        \includegraphics[width=1\linewidth]{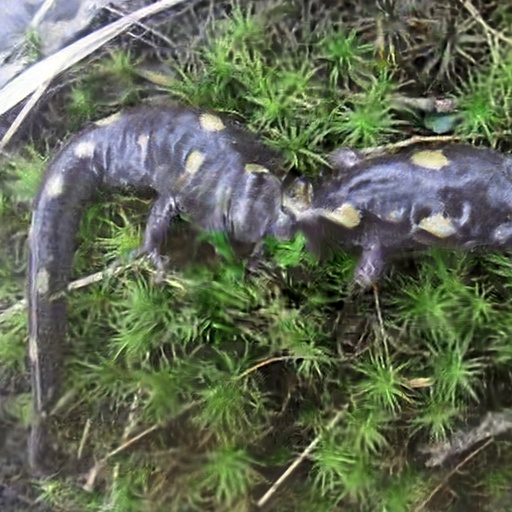}
        \includegraphics[width=1\linewidth]{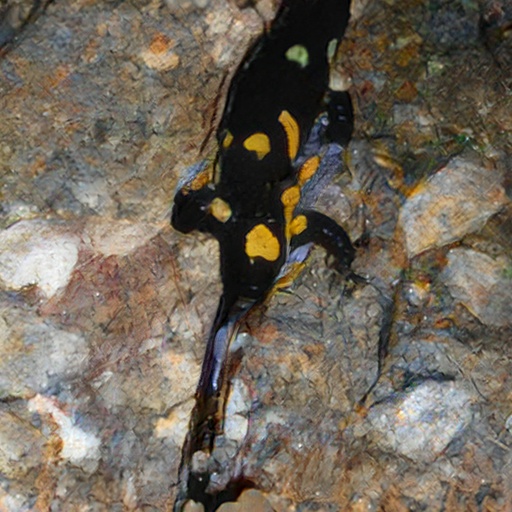}
        \includegraphics[width=1\linewidth]{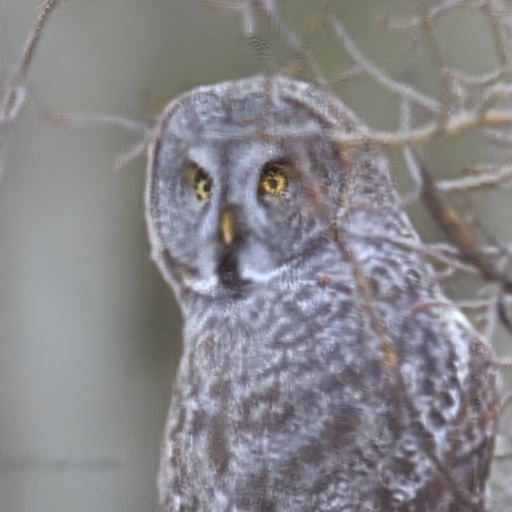}
    \end{minipage}
    \begin{minipage}{0.16\linewidth}
        \begin{center}
            GT
        \end{center}
        \includegraphics[width=1\linewidth]{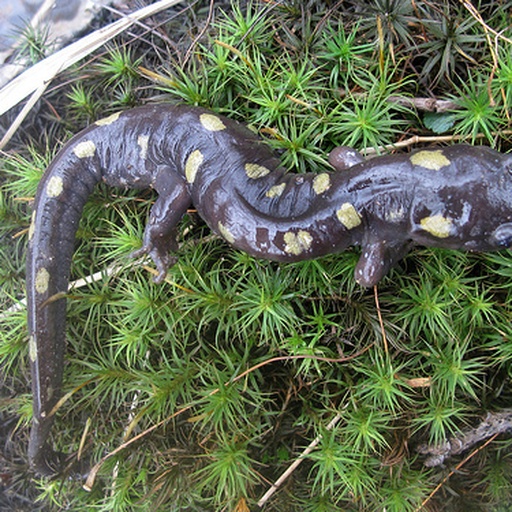}
        \includegraphics[width=1\linewidth]{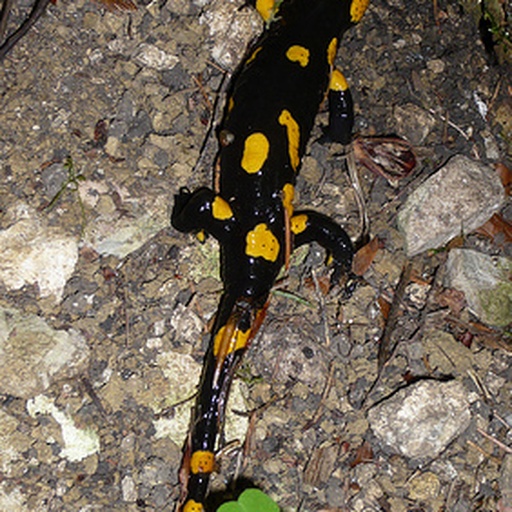}
        \includegraphics[width=1\linewidth]{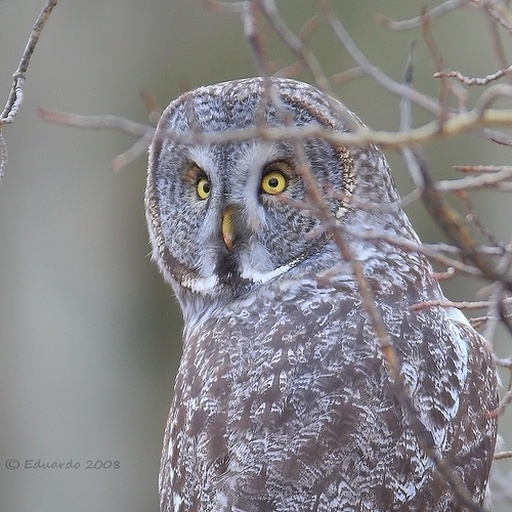}
    \end{minipage}
    \caption{Image Restoration with images from the ImageNet valid set. StyleGAN-XL pre-trained on ImageNet is used. Our method achieves more natural transitions, more realistic colors and sharper details. Zoom in for better visualization. }
    \label{fig:imagenet}
\end{figure*}
\subsubsection{Qualitative Results}
Figure \ref{fig:imagenet} presents a qualitative comparison against other GAN-inversion methods \cite{menon2020pulse, pan2021exploiting, roich2021pivotal} on ImageNet dataset. Our method achieves superior results for all tasks, i.e. image inpainting, image colorization and image super-resolution. Due to the lack of finetuning stage, PULSE fails to generalize to complex real-world images. It can only produce the general shape, but could not fill in the missing information. At the same time, since the latent vector and the generator are simultaneously optimized and finetuned, it is difficult for DGP to find a faithful output. Finetuning with incorrect latent vectors disrupts the generative capacity of the pre-trained generator which result in poor visual quality. Compared to PTI, our method is capable of recovering realistic fine details (\textit{e.g.} smoother contours for inpainting, truer colors for colorization and sharper details for SR). 

\begin{table}[t]
    \caption{Quantitative comparison against DGP and PTI on image inpainting, image colorization and image super-resolution. We use StyleGAN-XL pre-trained on ImageNet, and invert images from the ImageNet valid set.}
    \label{tab:imagenet}
    \centering
    {\begin{tabular}{lcccc}
    \toprule
       Task & Measure  &  DGP & PTI & Ours \\
   \midrule
       \multirow{2}*{Inpainting} & LPIPS $\downarrow$  & 0.2612 & 0.1219 & \textbf{0.1164} \\
       &FID $\downarrow$ & 166.24 & 54.62 & \textbf{44.73} \\
    \midrule
        \multirow{2}*{Colorization} & LPIPS $\downarrow$  & 0.3105 & 0.1994 & \textbf{0.1914}\\
       &FID $\downarrow$ & 78.32 & 47.51 & \textbf{45.90}\\
    \midrule
        \multirow{2}*{SR} & LPIPS $\downarrow$ & 0.2726 & 0.2429 & \textbf{0.2390}\\
       &NIQE $\downarrow$ & 8.4044 & 5.5659 & \textbf{5.3065}\\
   \bottomrule
    \end{tabular}}
\end{table}
\subsubsection{Quantitative Results} 
Following \cite{wu2021towards, zeng2021cr}, we use LPIPS~\cite{zhang2018unreasonable}, FID~\cite{heusel2017gans} for image inpainting and image colorization. Following \cite{pan2021exploiting}, LPIPS and NIQE~\cite{mittal2012making} are used for SR. Also, we find these metrics are close to visual perception. As shown in Table \ref{tab:imagenet}, the quantitative results align with our qualitative results as we achieve the best score for all metrics. 

\subsubsection{Comparison with DGP-BigGAN}
\begin{figure}[t]
    \centering
    \begin{minipage}[b]{0.3\linewidth}
    {
    \begin{center}
        Input
    \end{center}
    \includegraphics[width=1\linewidth]{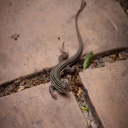}
    \includegraphics[width=1\linewidth]{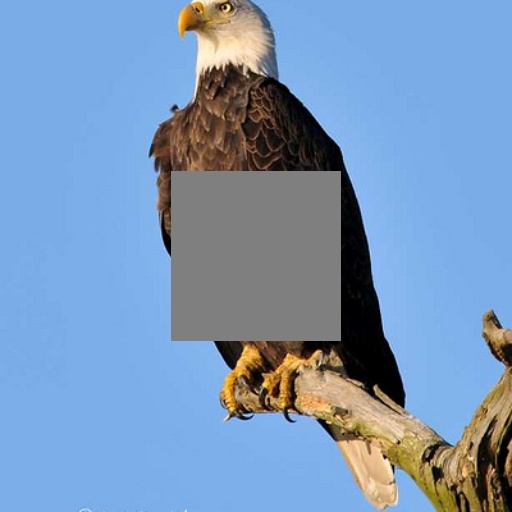}
    \includegraphics[width=1\linewidth]{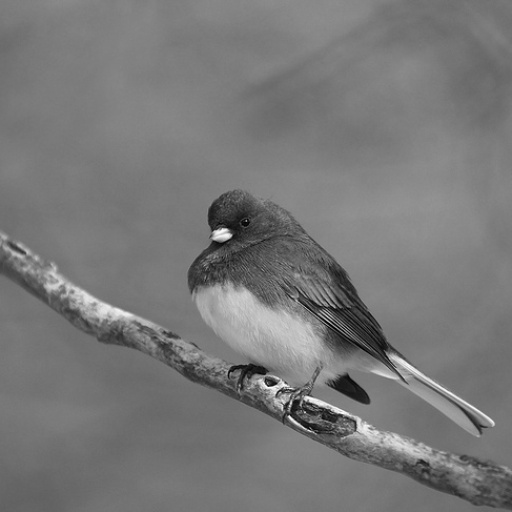}
    }
    \end{minipage}
    \begin{minipage}[b]{0.3\linewidth}
    {
    \begin{center}
        DGP
    \end{center}
    \includegraphics[width=1\linewidth]{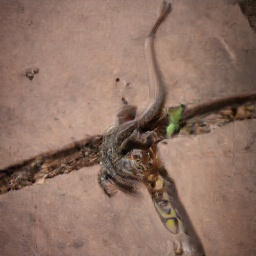}
    \includegraphics[width=1\linewidth]{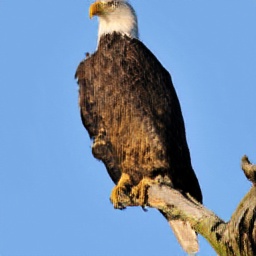}
    \includegraphics[width=1\linewidth]{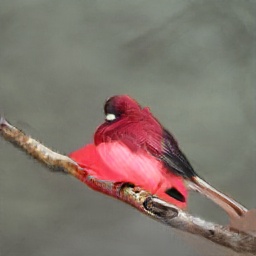}
    }
    \end{minipage}
    \begin{minipage}[b]{0.3\linewidth}
    {
    \begin{center}
        Ours
    \end{center}
    \includegraphics[width=1\linewidth]{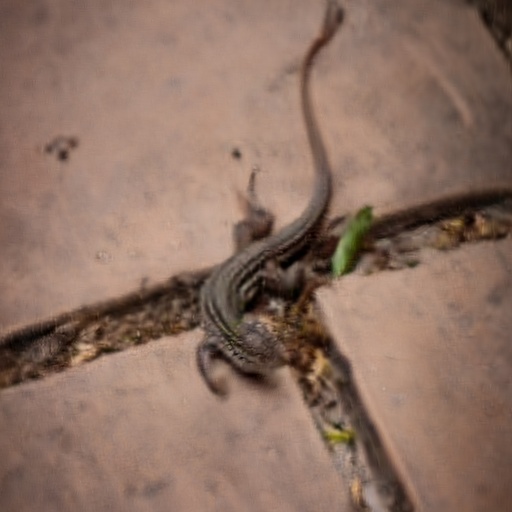}
    \includegraphics[width=1\linewidth]{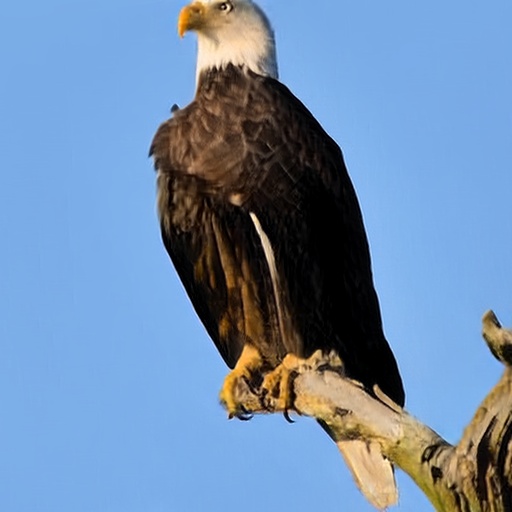}
    \includegraphics[width=1\linewidth]{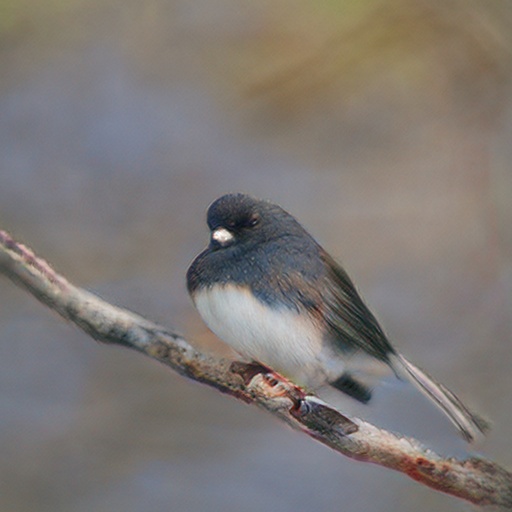}
    }
    \end{minipage}
    \caption{Qualitative comparison between DGP with BigGAN pre-trained on ImageNet and CRI with StyleGAN-XL pre-trained on ImageNet. Zoom in for better visualization. }
    \label{fig:biggan}
\end{figure}

We compare the results of DGP with BigGAN-ImageNet and the results of our methods. Quantitative and Qualitative results are shown in Table \ref{tab:biggan} and Figure \ref{fig:biggan}. Since BigGAN only supports the resolution of $256^2$, we resize the output of DGP+BigGAN to $512^2$ for comparison. For SR, the input of both DGP and CRI are set to $128^2$. Our CRI+StyleGAN-XL outperforms DGP+BigGAN on image inpainting, colorization and SR. 

\begin{table}[t]
    \caption{Quantitative comparison between DGP with BigGAN pre-trained on ImageNet and CRI with StyleGAN-XL pre-trained on ImageNet. }
    \label{tab:biggan}
    \centering
    \begin{tabular}{lccccc}
    \toprule
       Task & Measure  &  DGP & Ours \\
   \midrule
       \multirow{2}*{Inpainting} & LPIPS $\downarrow$  & 0.1316 & \textbf{0.1211} \\
       &FID $\downarrow$ & 116.12 & \textbf{80.20} \\
    \midrule
        \multirow{2}*{Colorization} & LPIPS $\downarrow$  & 0.2256 & \textbf{0.1842} \\
       &FID $\downarrow$ & 189.42 & \textbf{92.48} \\
    \midrule
        \multirow{2}*{SR} & LPIPS $\downarrow$ & 0.2226  & \textbf{0.2224} \\
       &NIQE $\downarrow$ & 9.2102 & \textbf{5.5686}\\
   \bottomrule
    \end{tabular}
\end{table}

\begin{figure}[t]
    \centering
    \subfigure[BSD100]{
        \includegraphics[width=0.23\linewidth]{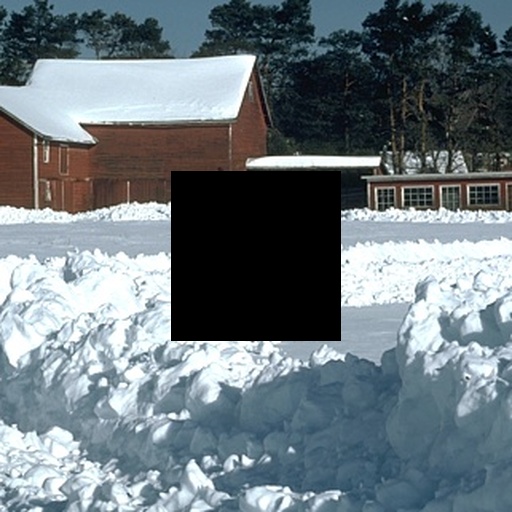}        \includegraphics[width=0.23\linewidth]{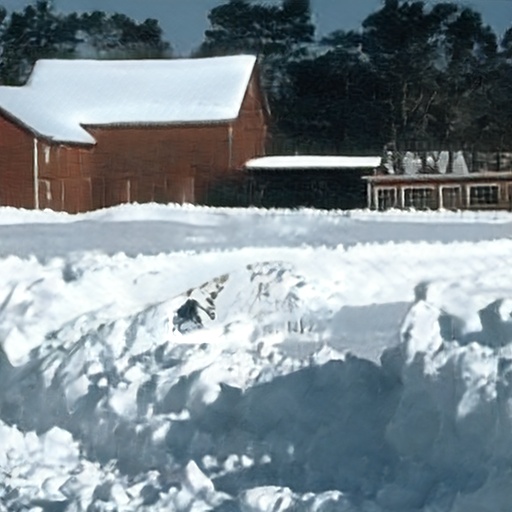}
    }
    \subfigure[DIV2K]{
	\includegraphics[width=0.23\linewidth]{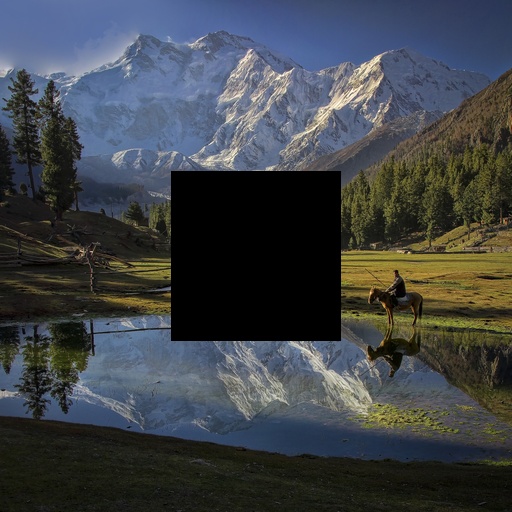}
    \includegraphics[width=0.23\linewidth]{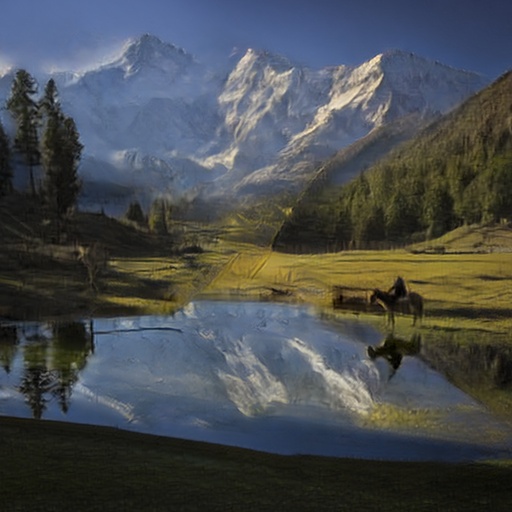}
}
    \caption{Evaluation of CRI on out-of-domain datasets, including (a) image from BSD100, (b) image from DIV2K.}
    \label{fig:ood}
\end{figure}

\subsection{Restoration on Face}
To illustrate our method is not limited by the GAN model or data, we conduct experiments on StyleGAN2~\cite{karras2020stylev2} pre-trained on FFHQ~\cite{karras2019stylev1}. The resolution of the generated images is $1024^2$. We invert 500 images from CelebA-HQ~\cite{karras2017progressive}. The scale factor of SR is set to 16. 
\subsubsection{Quantitative Results} 
For face restoration, we employ LPIPS and FID as evaluation metrics shown in Table \ref{tab:face}. The FID score is calculated between the inverted 500 images and the CelebA-HQ dataset (30000 images in total). Our method obtains the lowest LPIPS and FID for all tasks, indicating that our results are perceptually close to the ground truth and have a close distance to the real face distribution. 

\subsubsection{Qualitative Results}
\begin{figure*}[t]
    \centering
    \begin{minipage}{0.16\linewidth}
        \begin{center}
            Input
        \end{center}
        \includegraphics[width=1\linewidth]{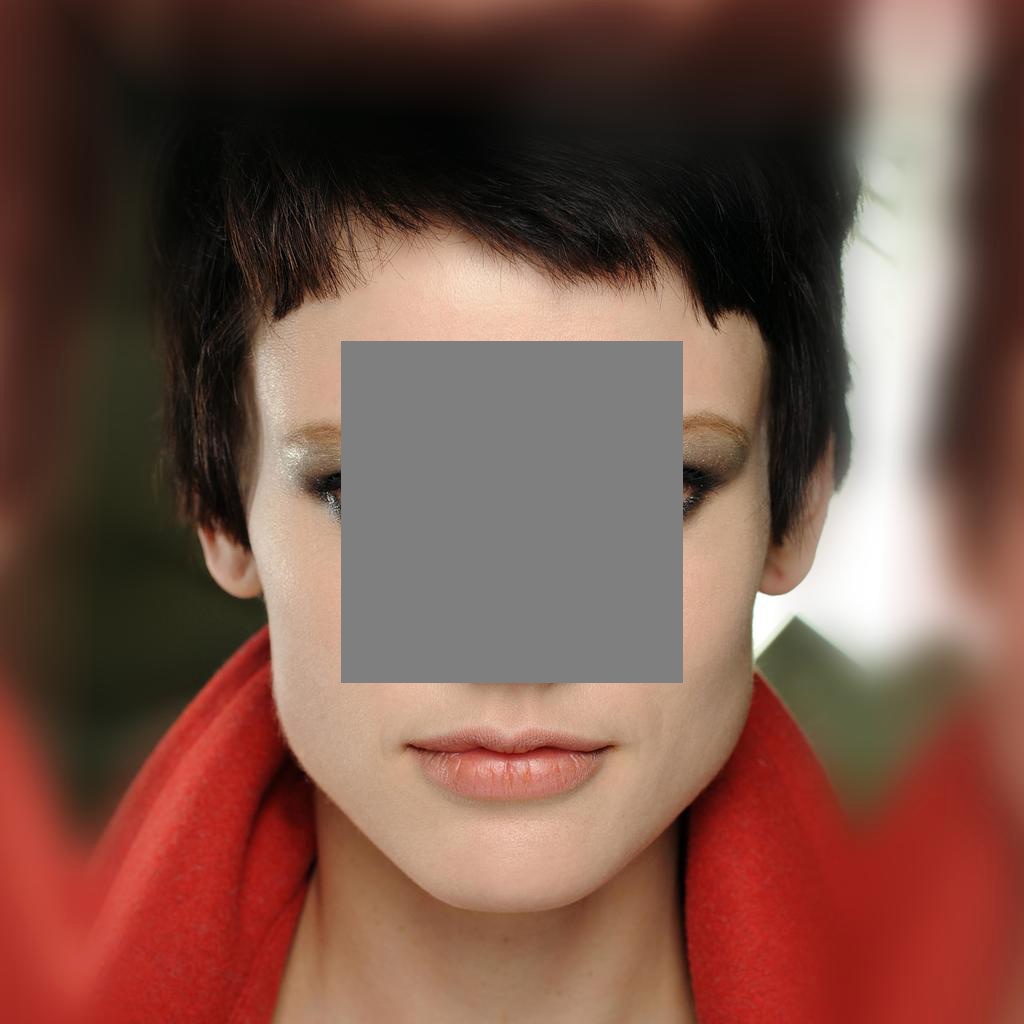}
        \includegraphics[width=1\linewidth]{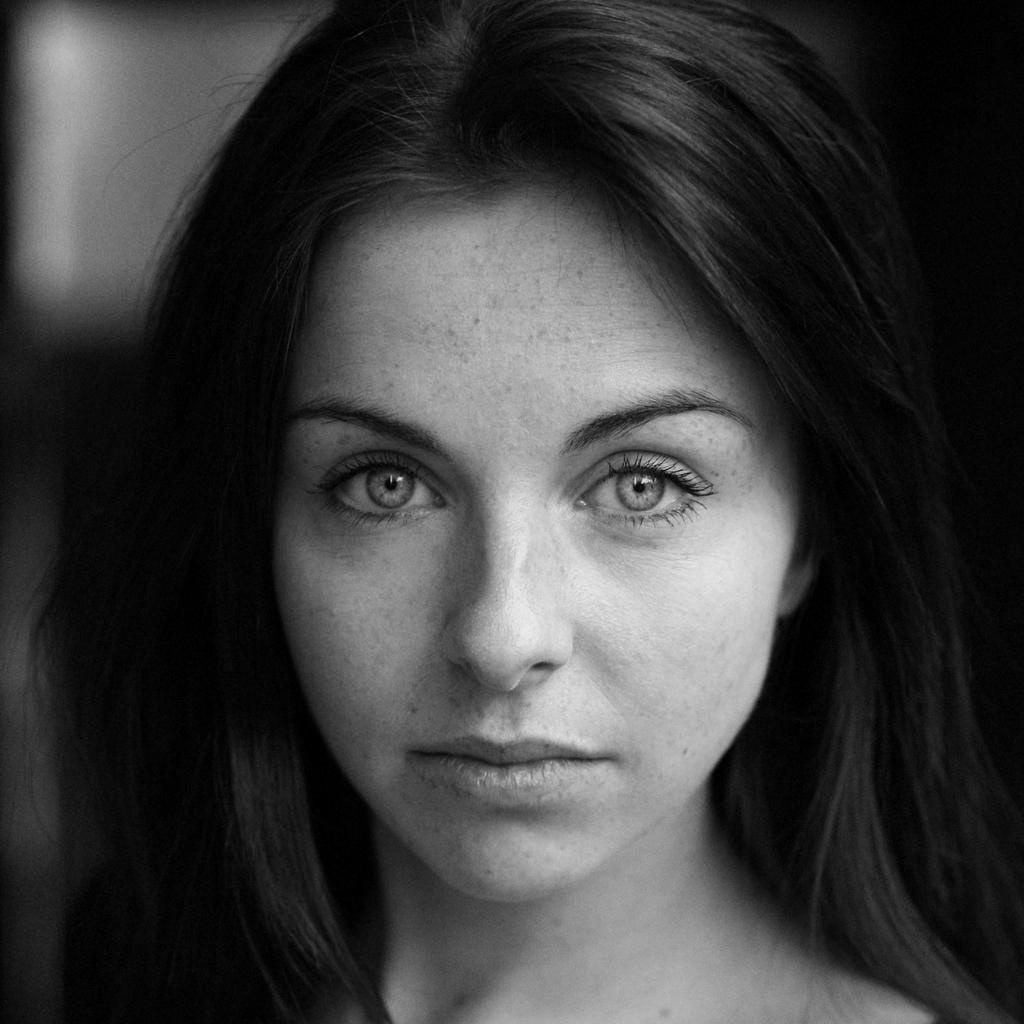}        \includegraphics[width=1\linewidth]{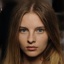}
    \end{minipage}
    \begin{minipage}{0.16\linewidth}
        \begin{center}
            PULSE
        \end{center}
        \includegraphics[width=1\linewidth]{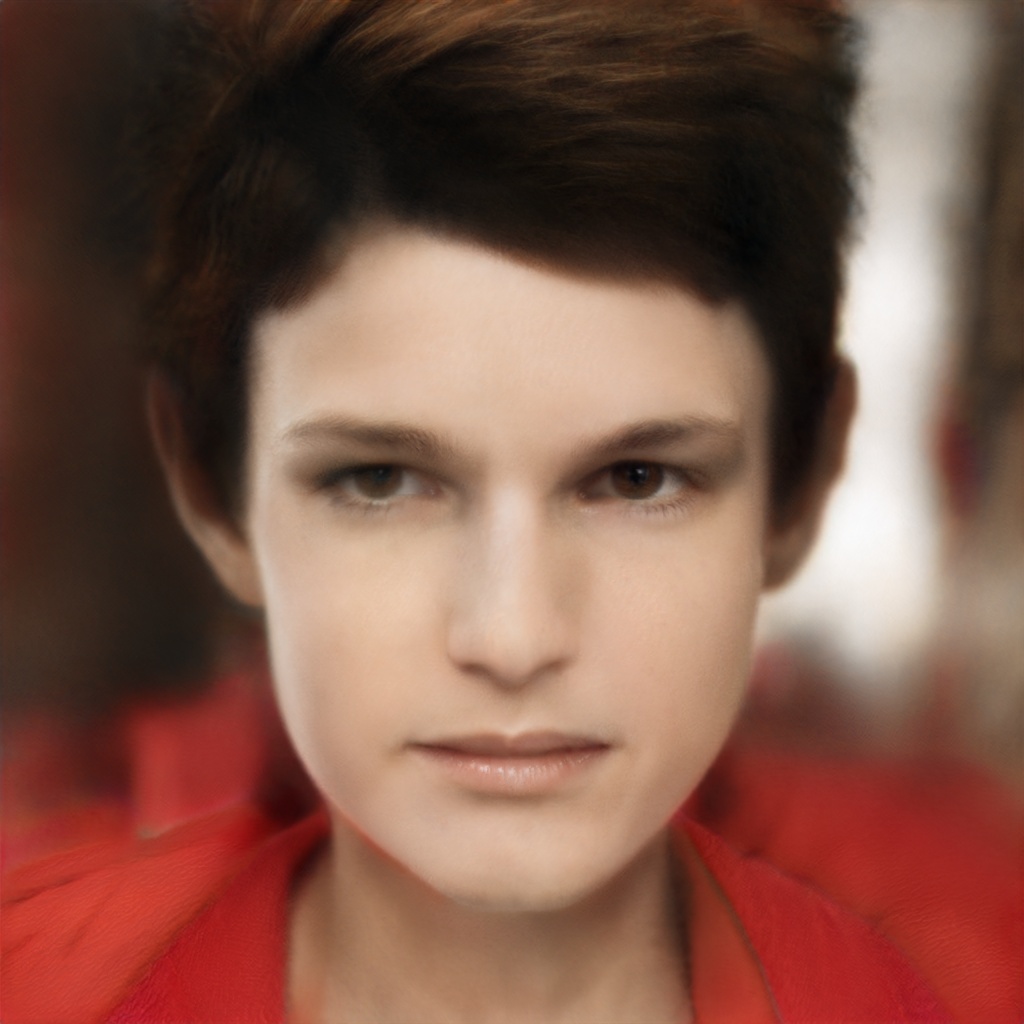}
        \includegraphics[width=1\linewidth]{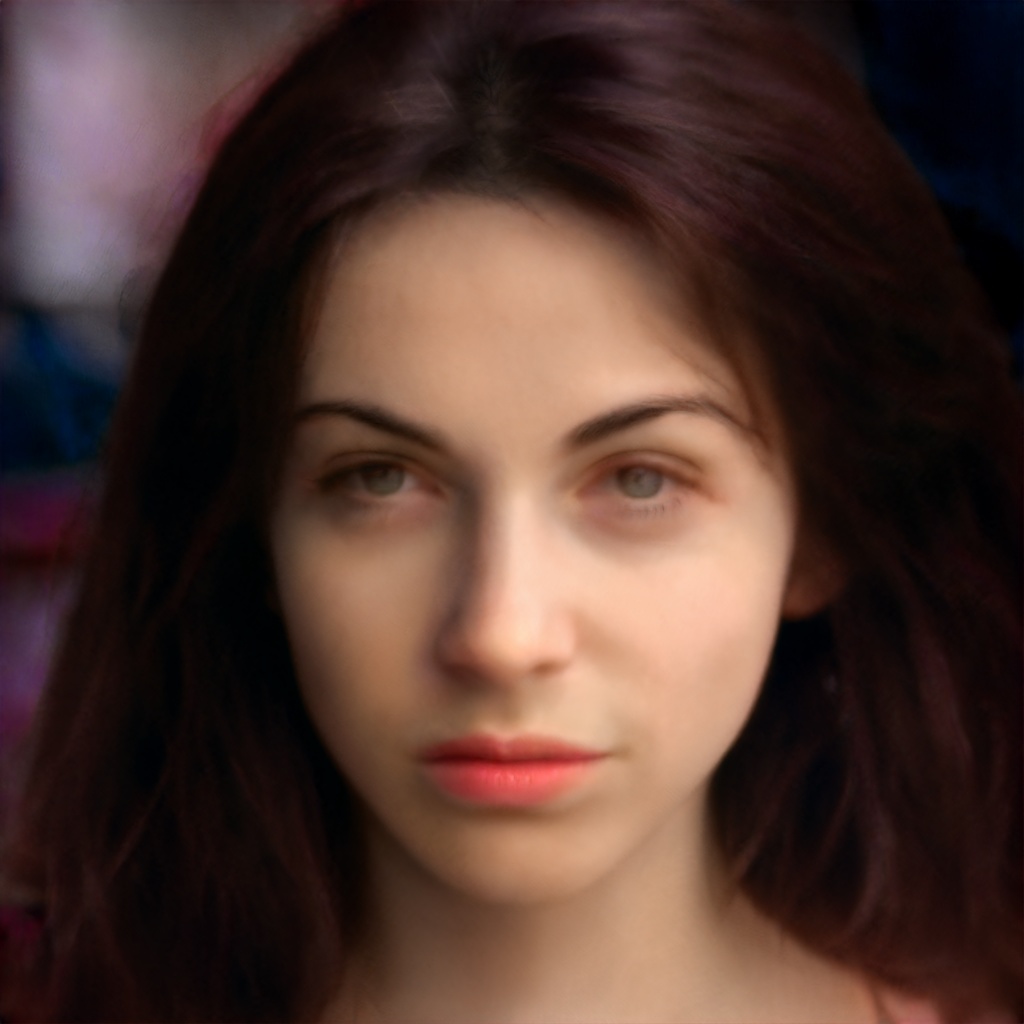}
        \includegraphics[width=1\linewidth]{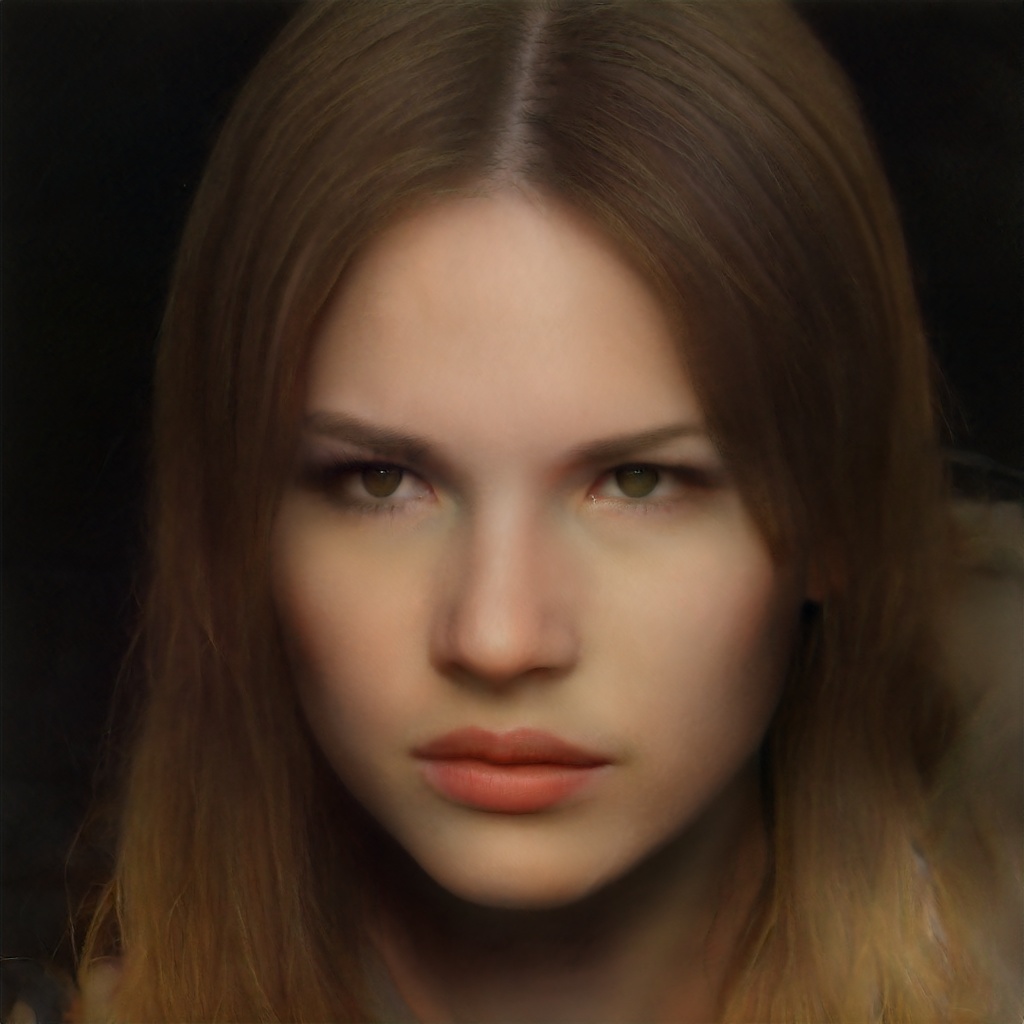}
    \end{minipage}
    \begin{minipage}{0.16\linewidth}
        \begin{center}
            DGP
        \end{center}
        \includegraphics[width=1\linewidth]{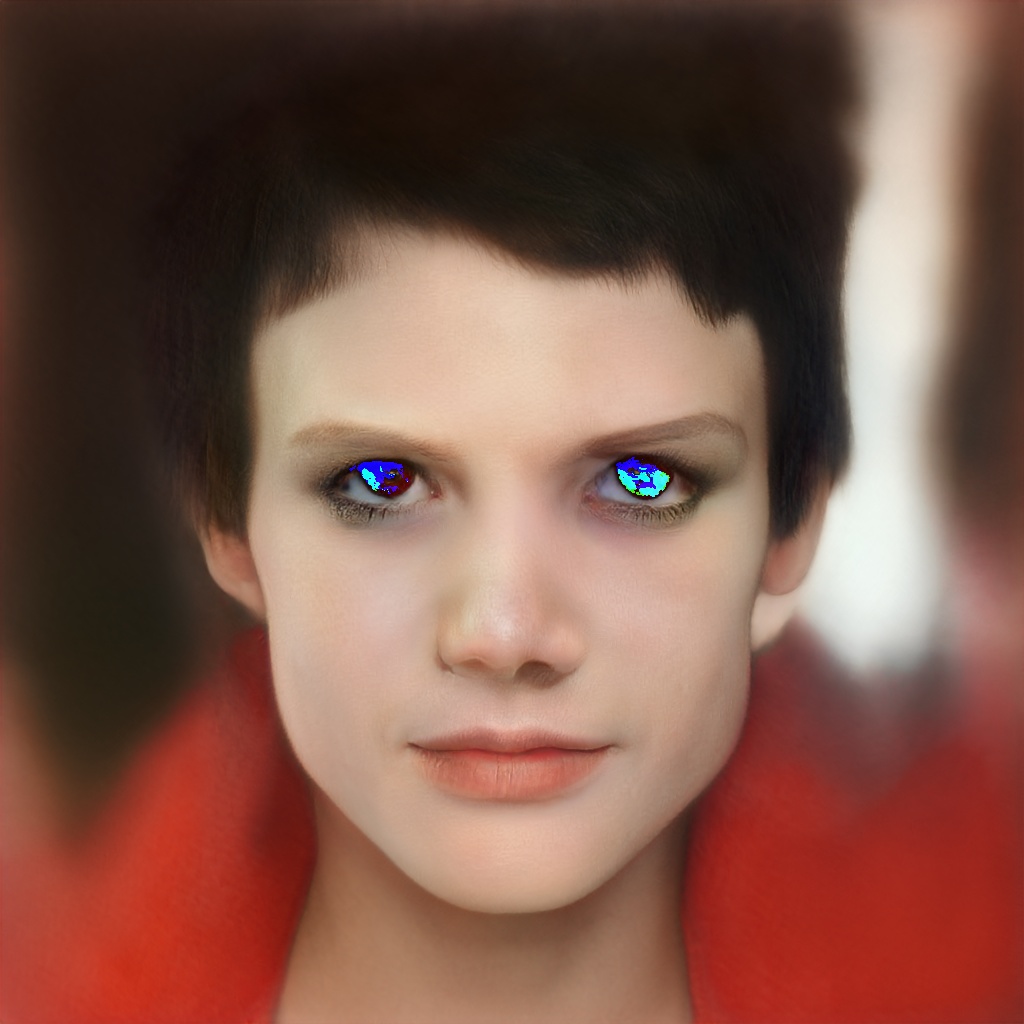}
        \includegraphics[width=1\linewidth]{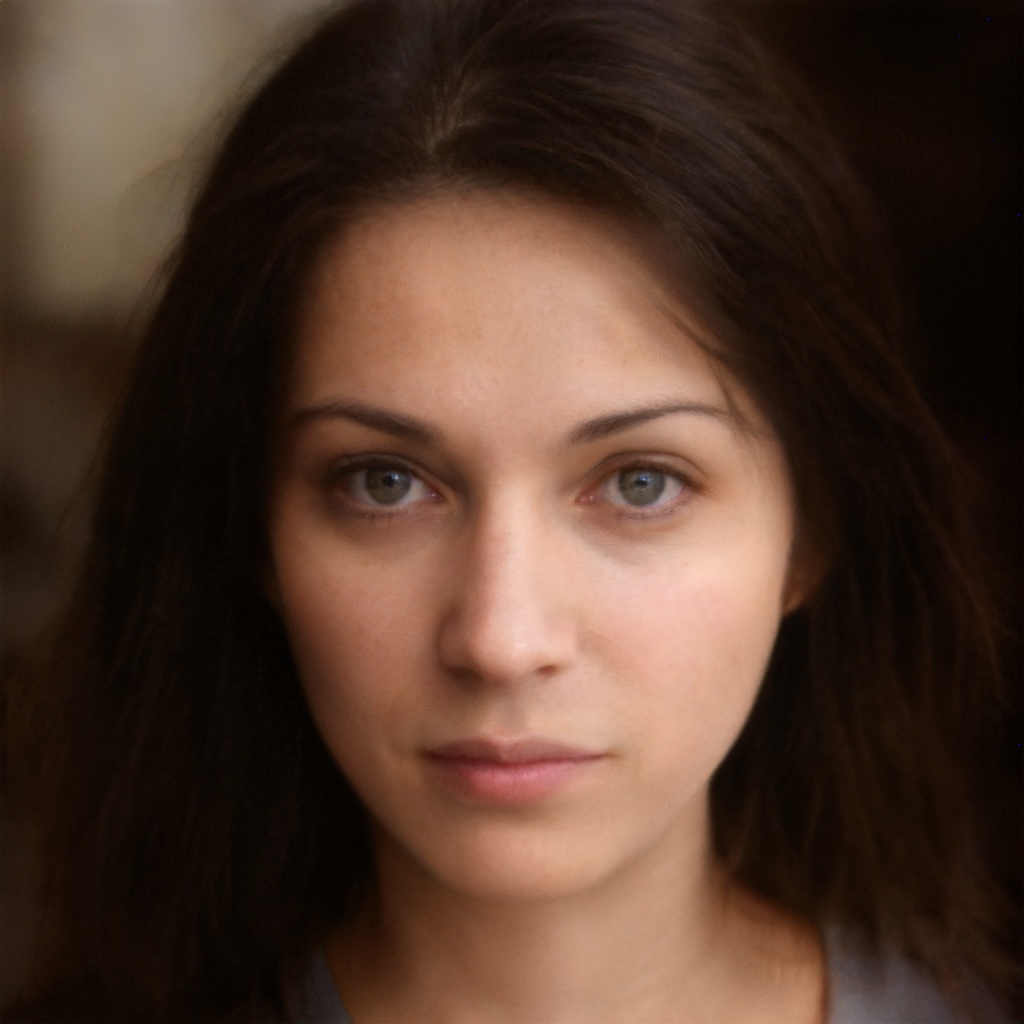}
        \includegraphics[width=1\linewidth]{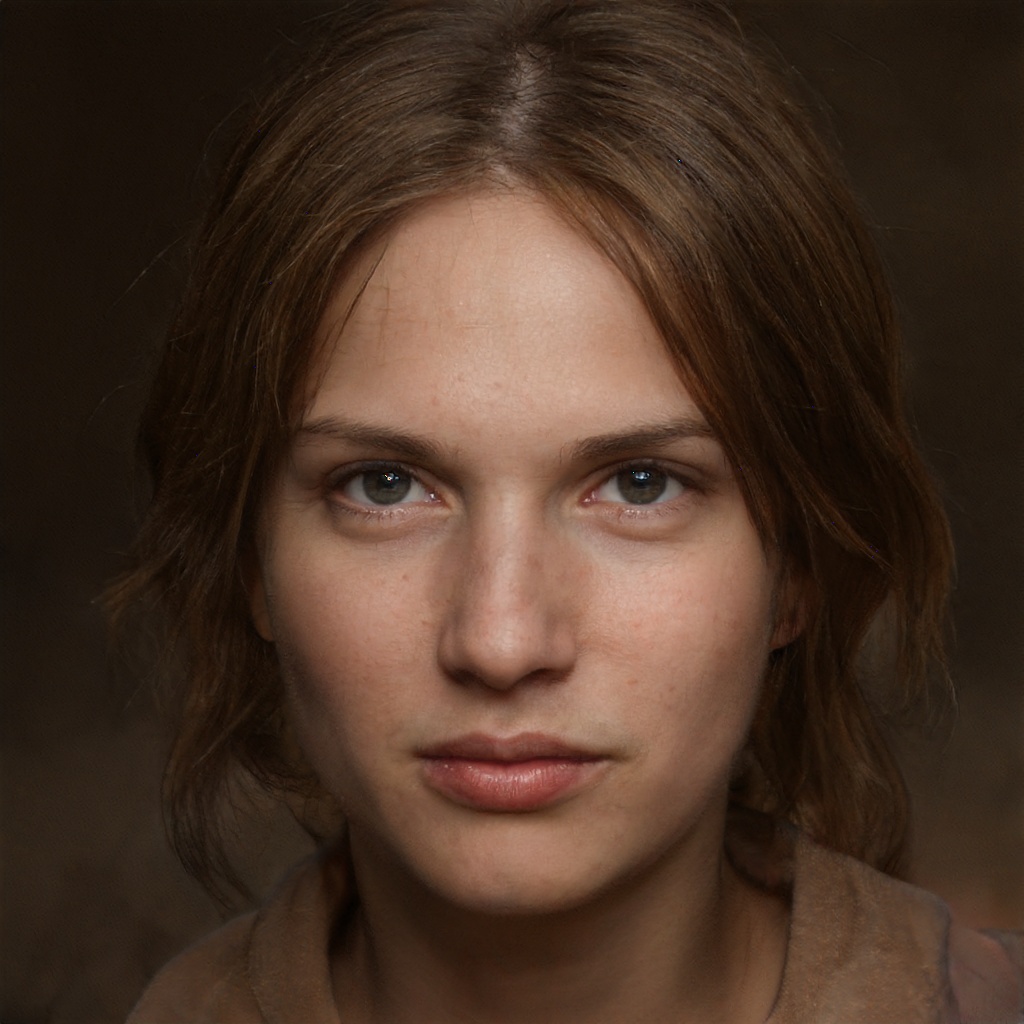}
    \end{minipage}
    \begin{minipage}{0.16\linewidth}
        \begin{center}
            PTI
        \end{center}
        \includegraphics[width=1\linewidth]{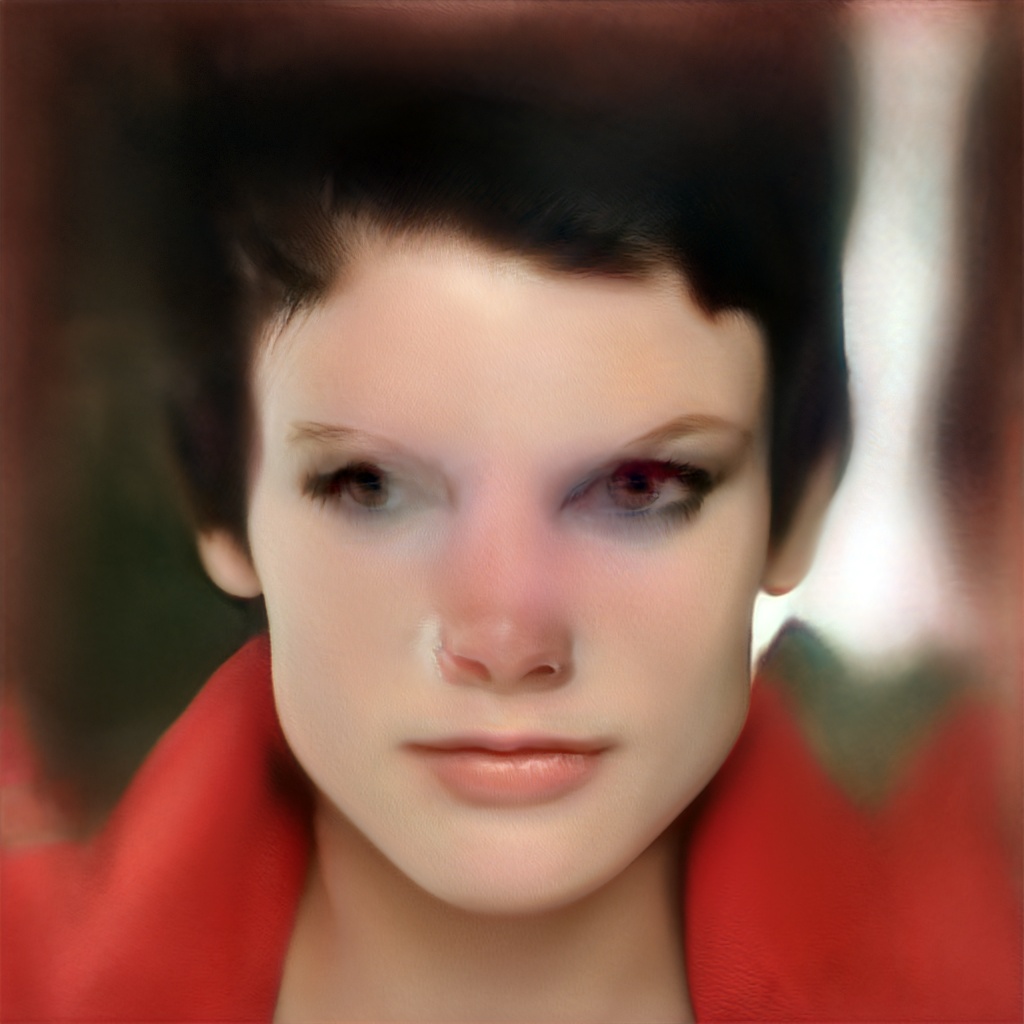}
        \includegraphics[width=1\linewidth]{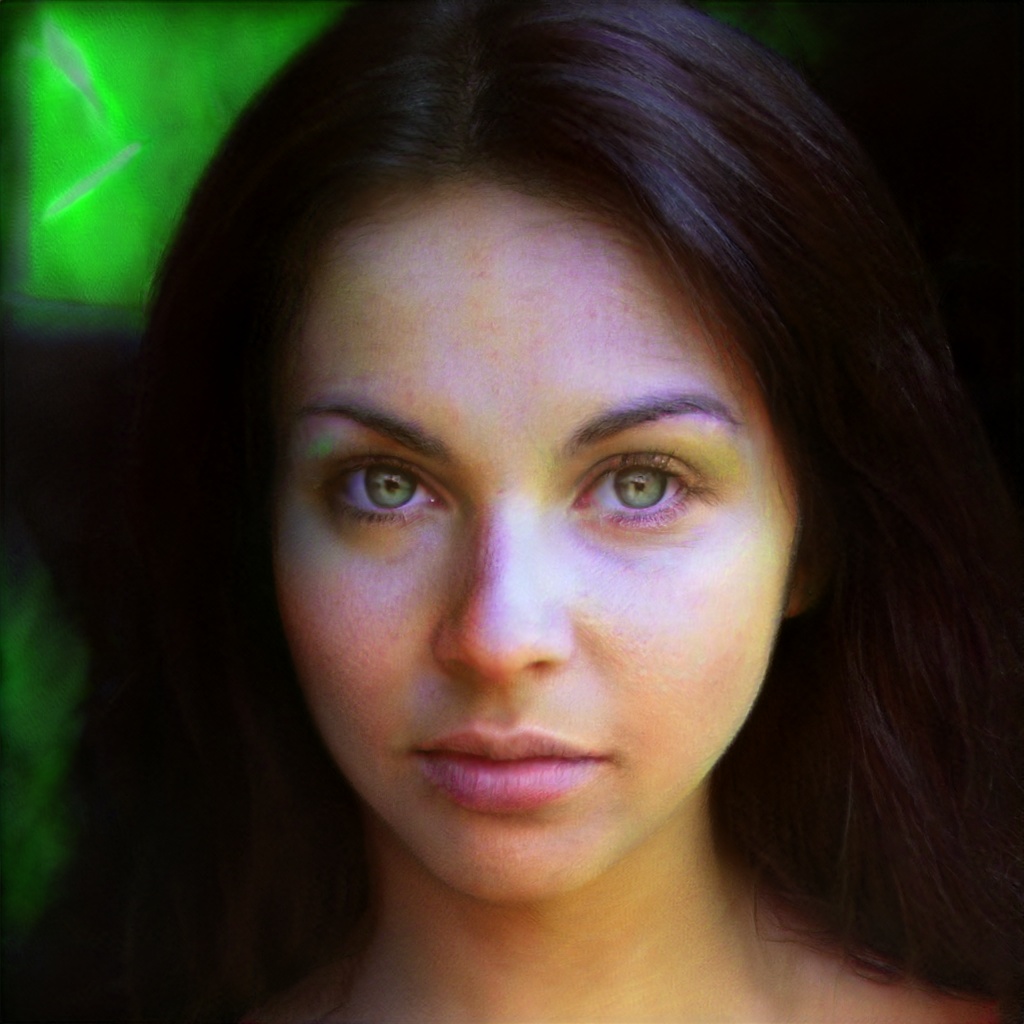}
        \includegraphics[width=1\linewidth]{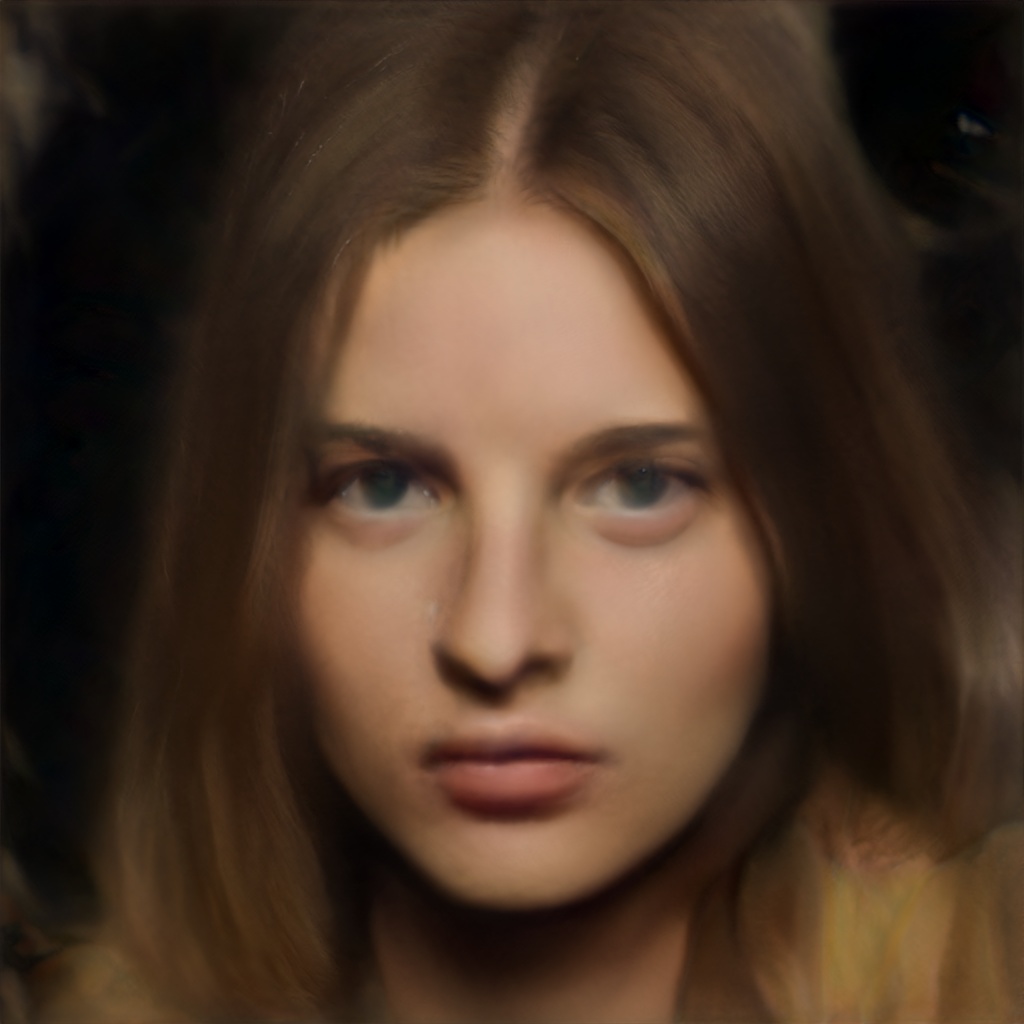}
    \end{minipage}
    \begin{minipage}{0.16\linewidth}
        \begin{center}
            Ours
        \end{center}
        \includegraphics[width=1\linewidth]{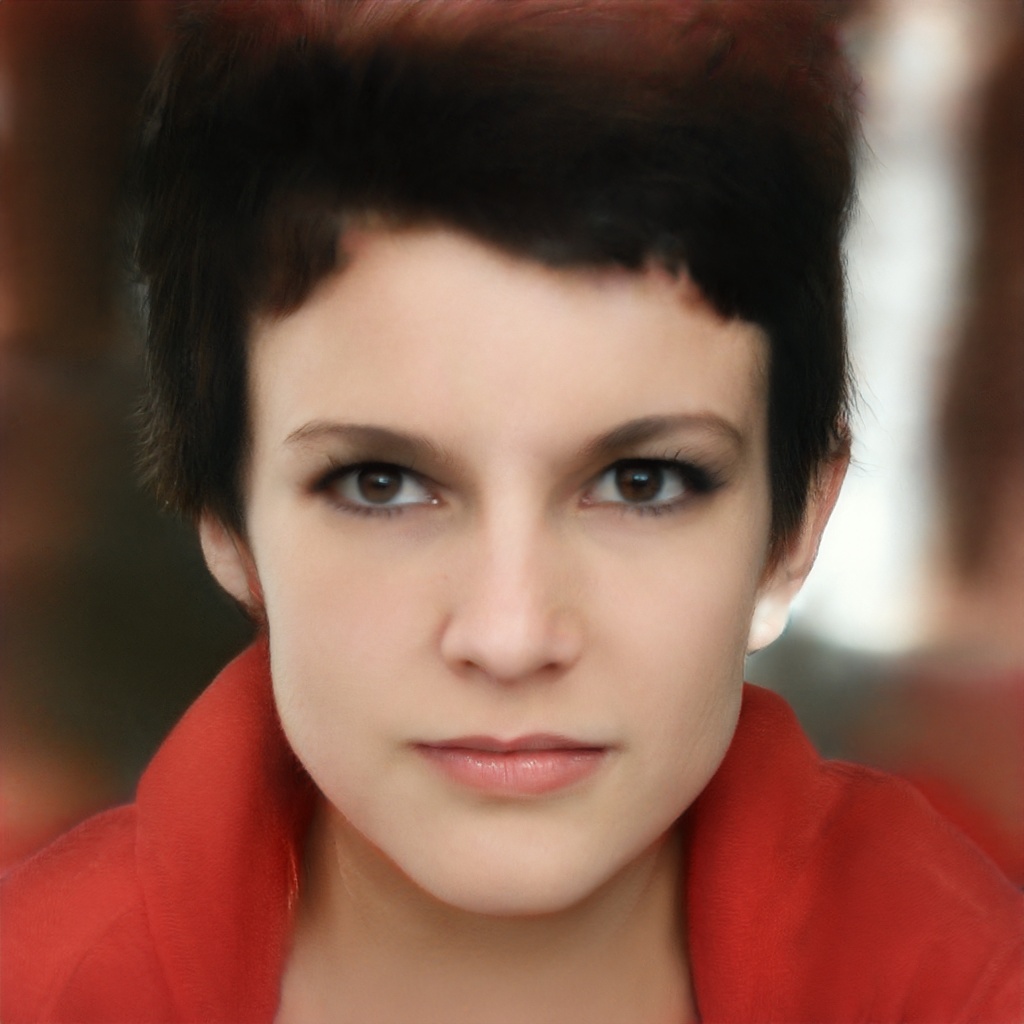}
        \includegraphics[width=1\linewidth]{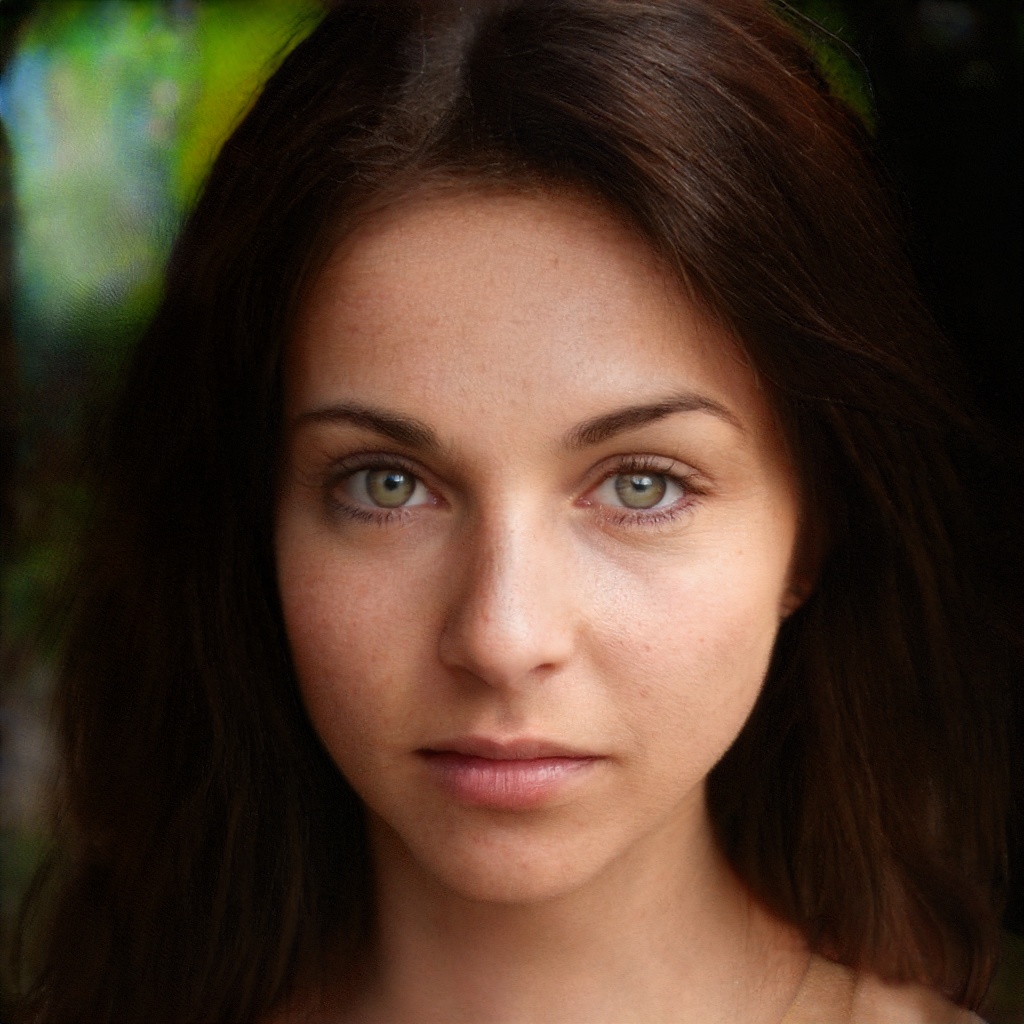}
        \includegraphics[width=1\linewidth]{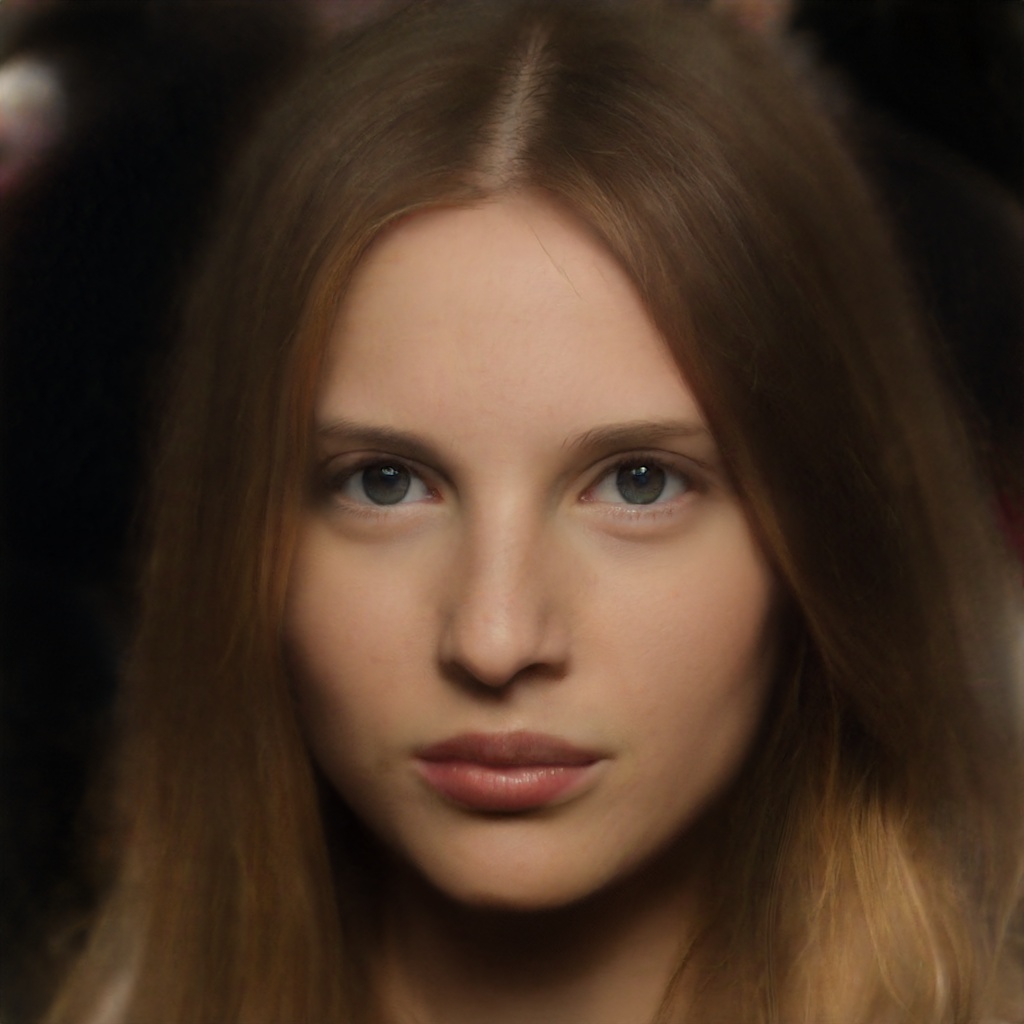}
    \end{minipage}
    \begin{minipage}{0.16\linewidth}
        \begin{center}
            GT
        \end{center}
        \includegraphics[width=1\linewidth]{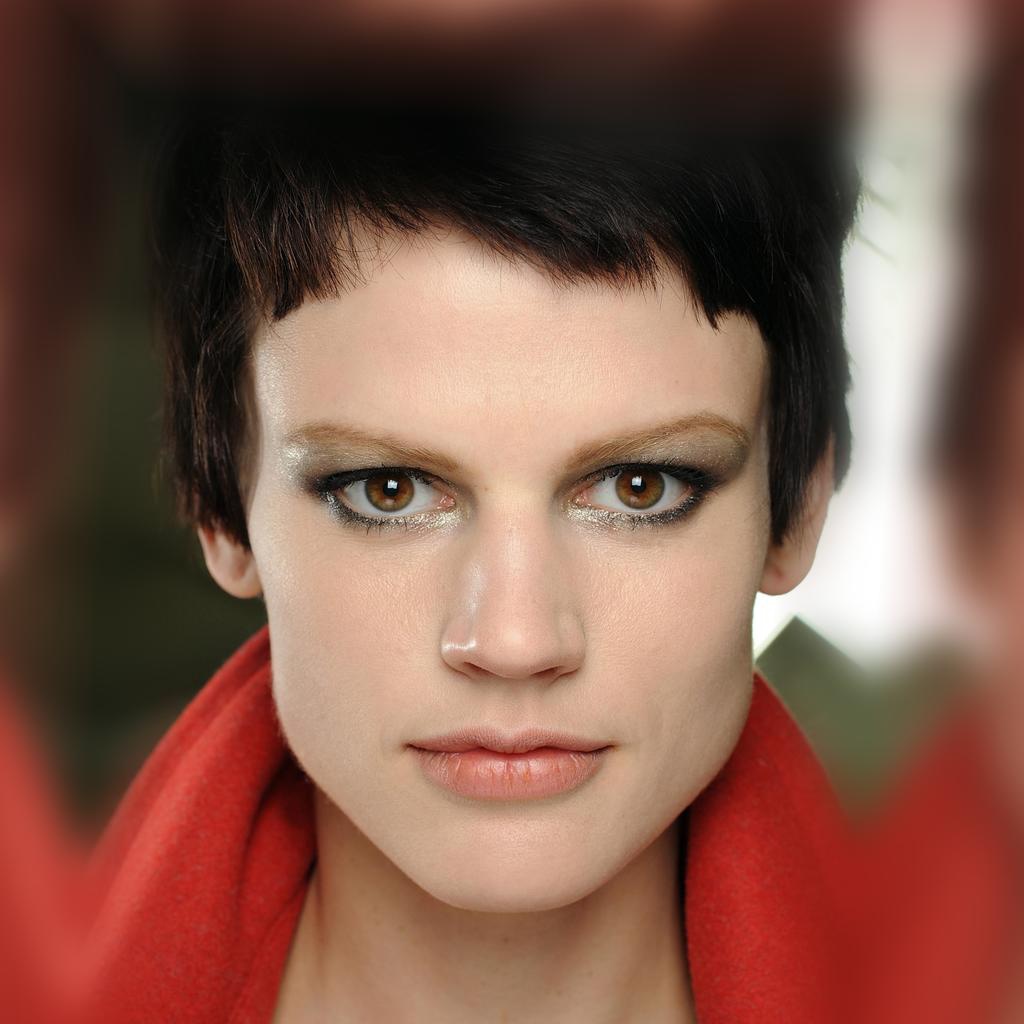}
        \includegraphics[width=1\linewidth]{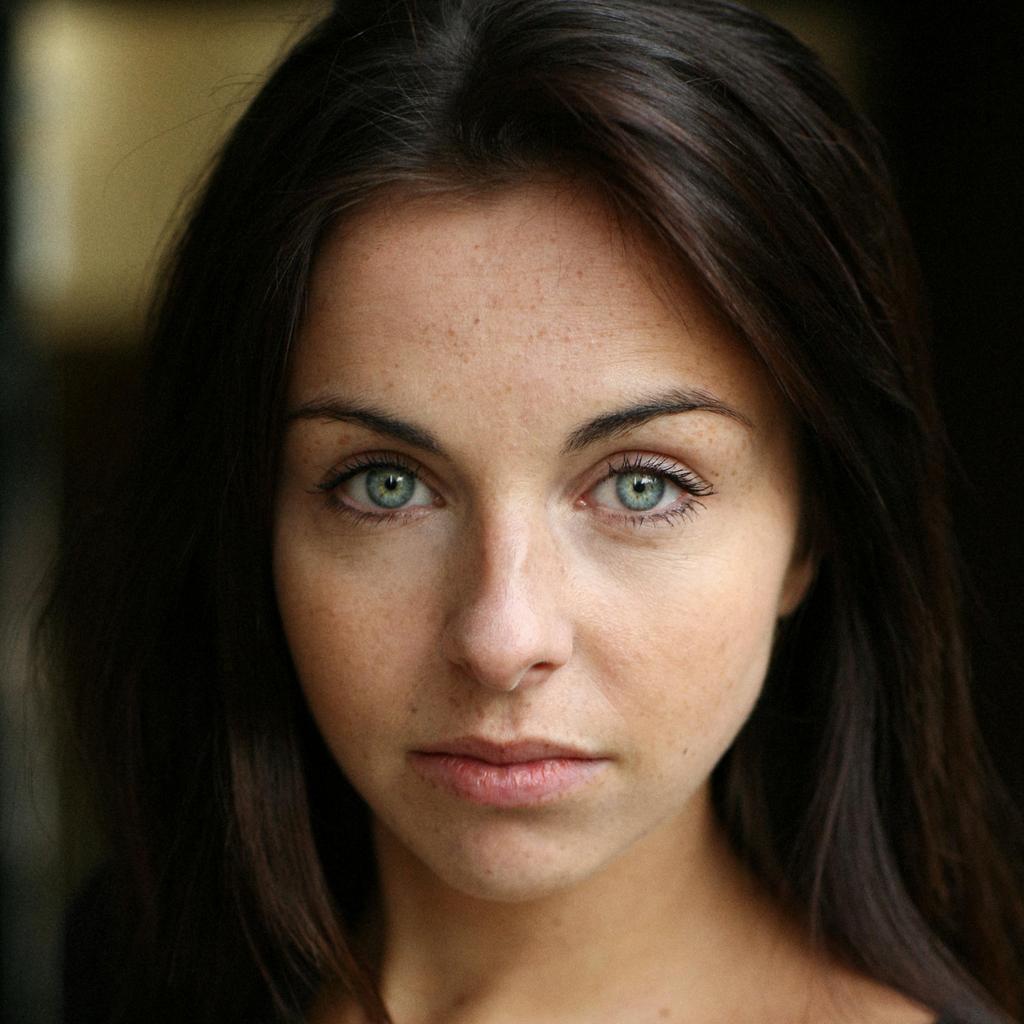}
        \includegraphics[width=1\linewidth]{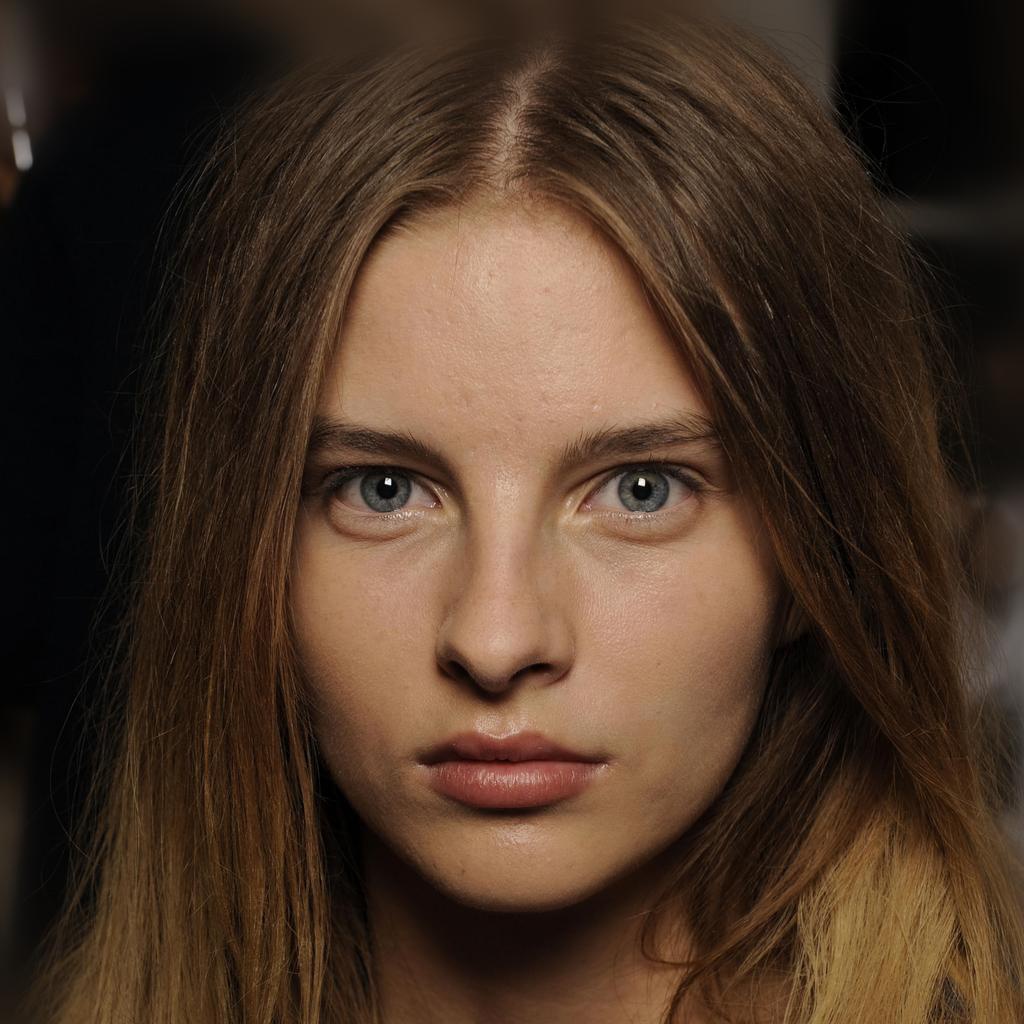}
    \end{minipage}
    \caption{Image Restoration with images from the CelebA-HQ dataset. StyleGAN2 pre-trained on FFHQ is used for all methods. Our method guaranteed similarity while generating high quality face images. Zoom in for better visualization.}
    \label{fig:celeba}
\end{figure*}

Qualitative results of each task are shown in Figure \ref{fig:celeba}. Our method outperforms other methods in terms of image quality and identity preservation. PULSE and DGP can produce results of good visual quality, but the similarity is relatively low. For instance, DGP-SR fails to preserve the hairstyle. Images generated by PTI are similar to the input in terms of known information. For example, the background of the image in the inpainting task and the structure in the colorization task are quite authentic. However, the overall view of the generated images is not natural since the inverted latent vectors beyond the manifold of natural images. Compared to PTI, our results are more natural and generate sharper details. 
\begin{table}[t]
    \caption{Quantitative comparison against PULSE, DGP and PTI on image inpainting, image colorization and image super-resolution. We use StyleGANv2 pre-trained on FFHQ, and invert images from the CelebA-HQ.}
    \label{tab:face}
    \centering
    \resizebox{.95\linewidth}{!}{\begin{tabular}{lccccc}
    \toprule
       Task & Measure  &  DGP & PULSE & PTI & Ours \\
   \midrule
       \multirow{2}*{Inpainting} & LPIPS $\downarrow$  & 0.1311 & 0.1387 & \textit{0.1293} & \textbf{0.1279}\\
       &FID $\downarrow$ & 68.32 &59.20 & 53.48 & \textbf{48.00}\\
    \midrule
        \multirow{2}*{Colorization} & LPIPS $\downarrow$  & 0.1621 &0.1924 & 0.1990& \textbf{0.1505}\\
       &FID $\downarrow$ & 74.24 & 116.06 & 56.01 & \textbf{46.54}\\
    \midrule
        \multirow{2}*{SR} & LPIPS $\downarrow$ & 0.1481 & 0.1385 & \textit{0.1317} & \textbf{0.1297}\\
       &FID $\downarrow$ & 68.56 & 60.01 & 60.76 & \textbf{58.88}\\
   \bottomrule
    \end{tabular}}
\end{table}

\subsection{Robustness on Out-of-domain Images}
We also experiments with images from BSD100~\cite{martin2001database} dataset and DIV2K~\cite{agustsson2017ntire} dataset. Deit-M~\cite{touvron2021training} is used as a classifier to provide the class information (\textit{i.e.}, $c$ in Eq. \ref{eq:map}) for the input. As shown in Figure \ref{fig:ood}, our CRI can reconstruct the missing part of out-of-domain images. 
\begin{figure}[t]
    \centering
    \includegraphics[width=0.8\linewidth]{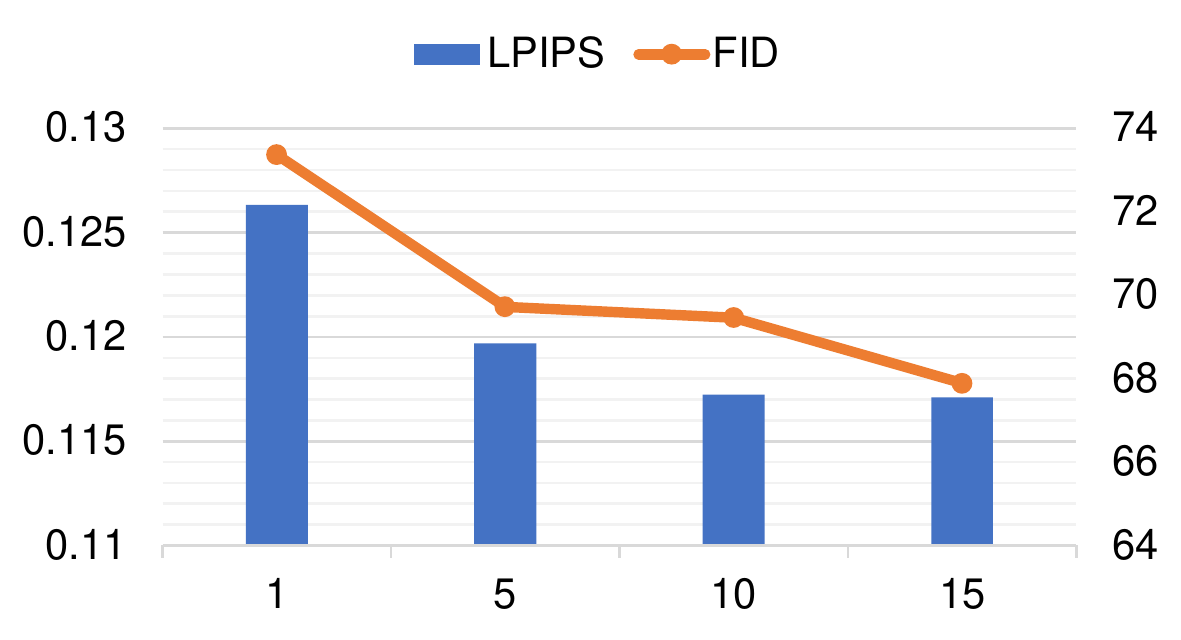}
    \caption{Comparison of numbers of the cluster.}
    \label{fig:centroids}
\end{figure}

\subsection{Ablation Study}
\paragraph{Effect of Centroids}
We show the effectiveness of the clustering centroids by adopting different cluster numbers. To exclude the effect of the random sampling of $\{z_j\}_{j=1}^M$, we cluster the same $\{z_j\}_{j=1}^M$ into $N = \{1, 5, 10, 15\}$ clusters respectively using KMeans. The results of colorization are presented in Figure \ref{fig:centroids}. The performance becomes progressively better as the number of centroids increases. However, the clustering time also becomes longer. Considering the performance and the clustering time trade-off, we choose $N=10$ as the default. 

\begin{figure}[t]
    \centering
    \begin{minipage}[b]{0.23\linewidth}
    {
    \begin{center}
        Input
    \end{center}
    \includegraphics[width=1\linewidth]{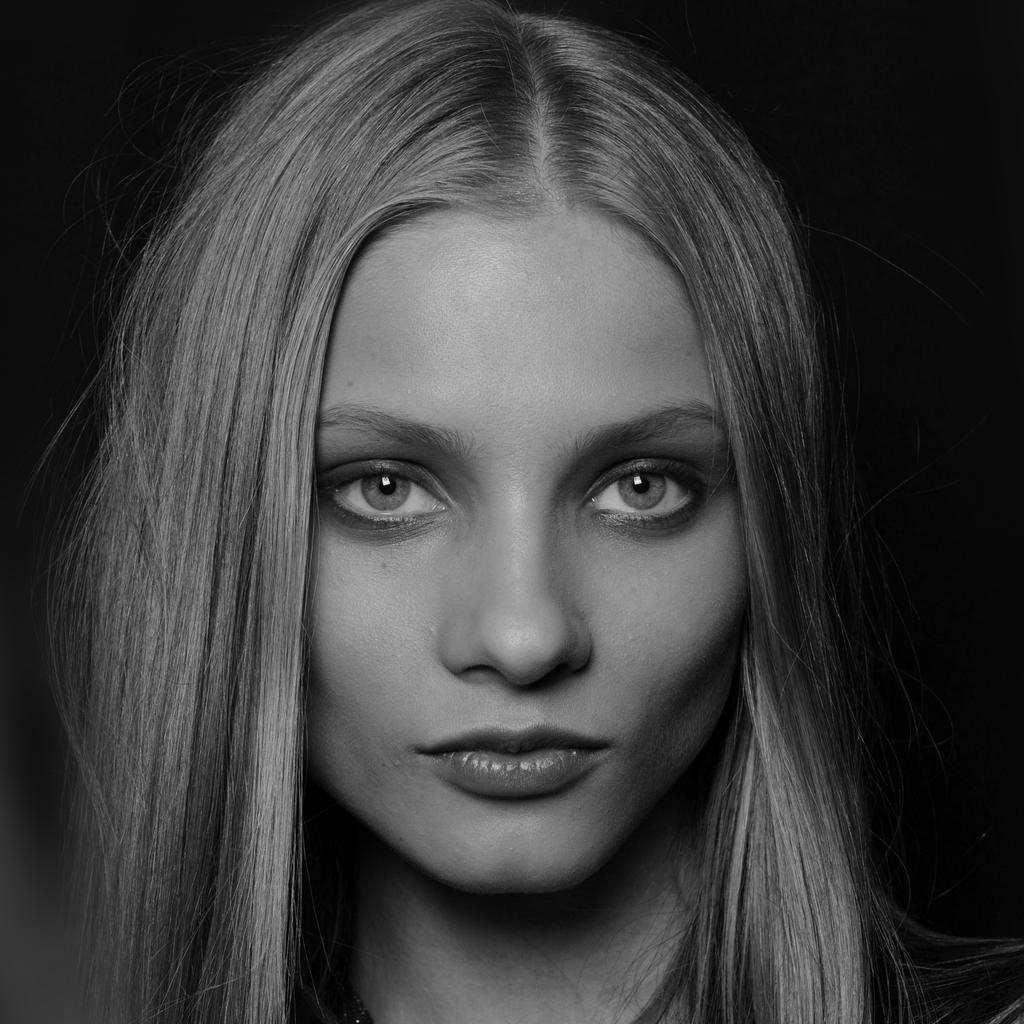}
    \includegraphics[width=1\linewidth]{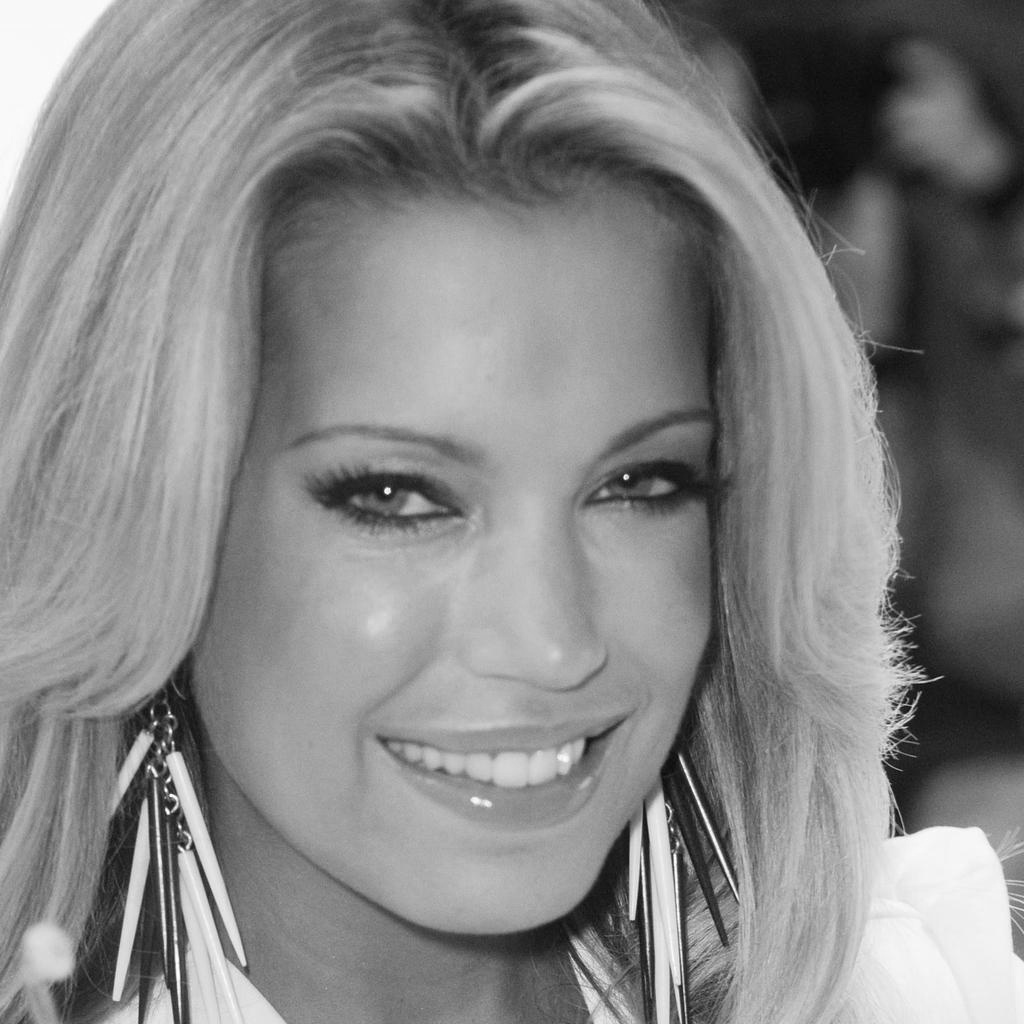}
    }
    \end{minipage}
    \begin{minipage}[b]{0.23\linewidth}
    {
    \begin{center}
        w/o Reg
    \end{center}
    \includegraphics[width=1\linewidth]{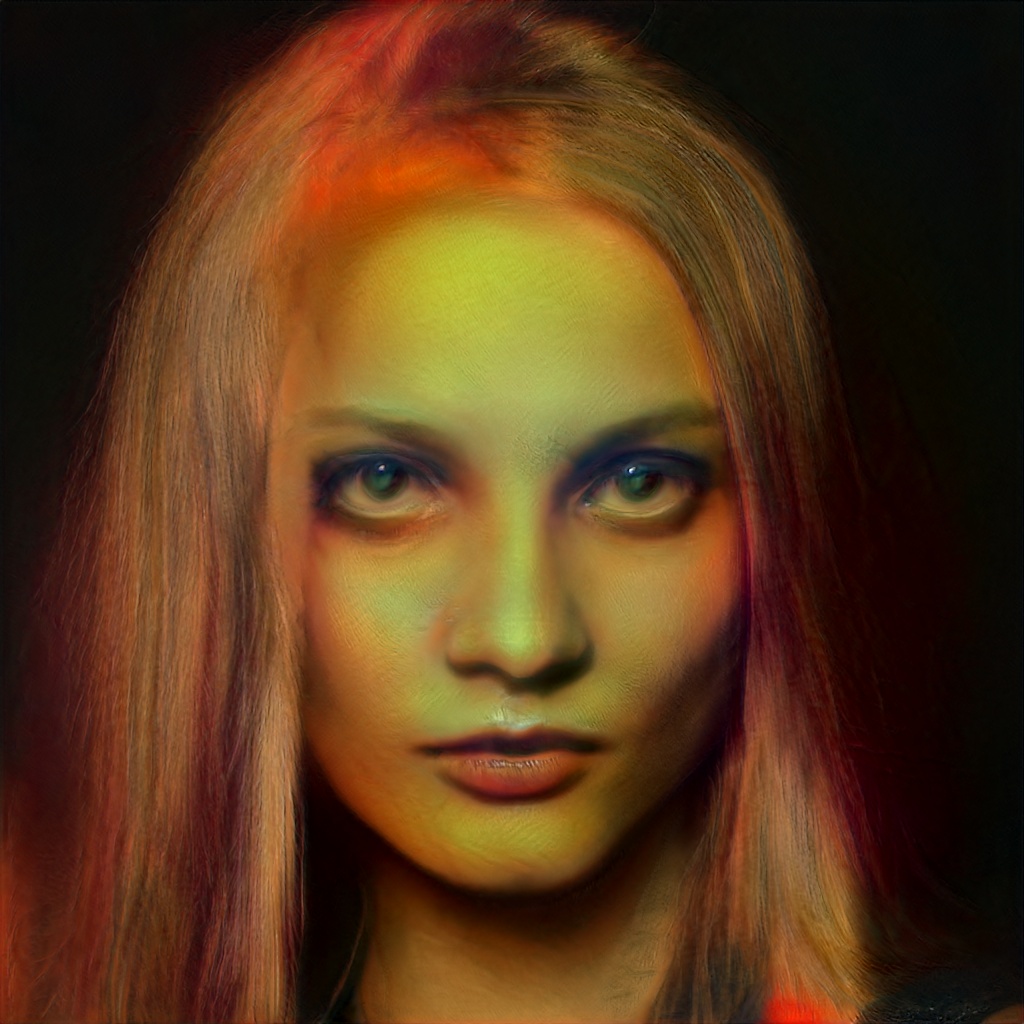}
    \includegraphics[width=1\linewidth]{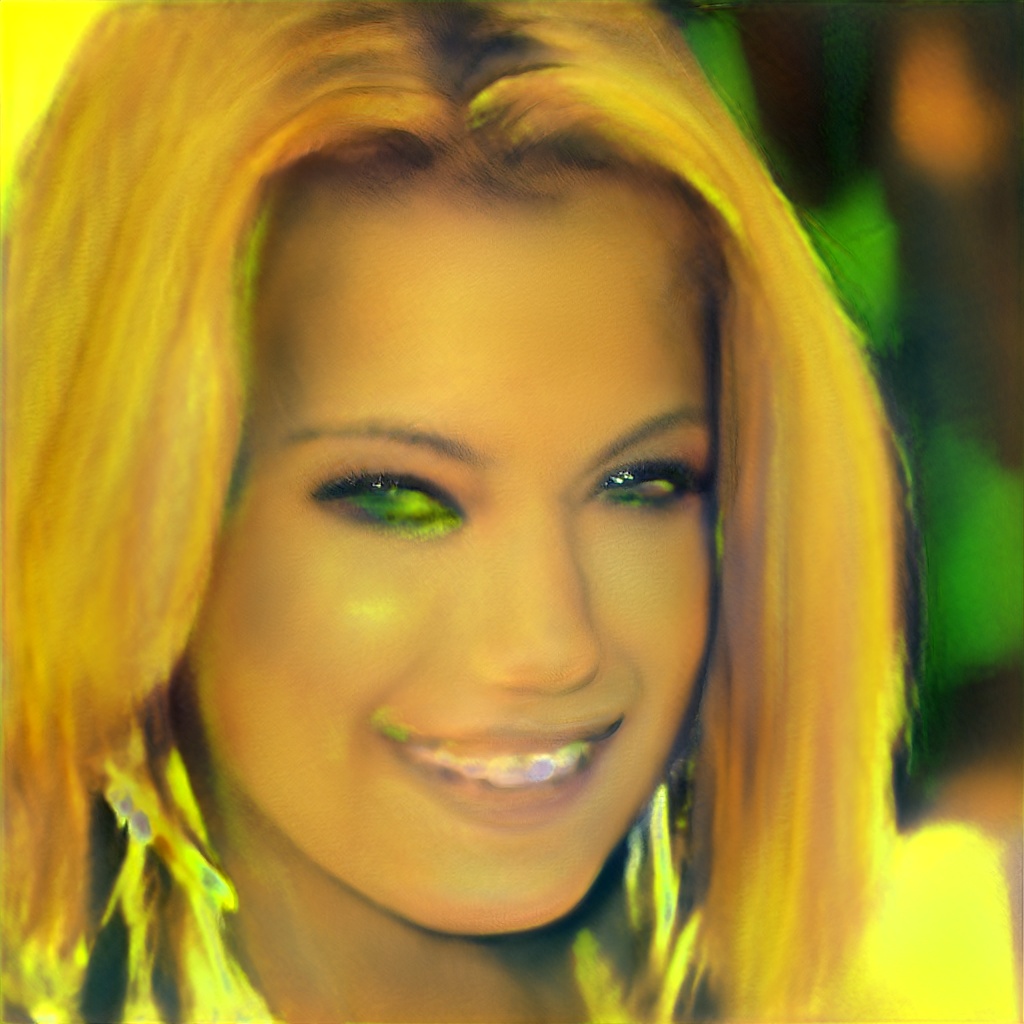}
    }
    \end{minipage}
    \begin{minipage}[b]{0.23\linewidth}
    {
    \begin{center}
        L1
    \end{center}
    \includegraphics[width=1\linewidth]{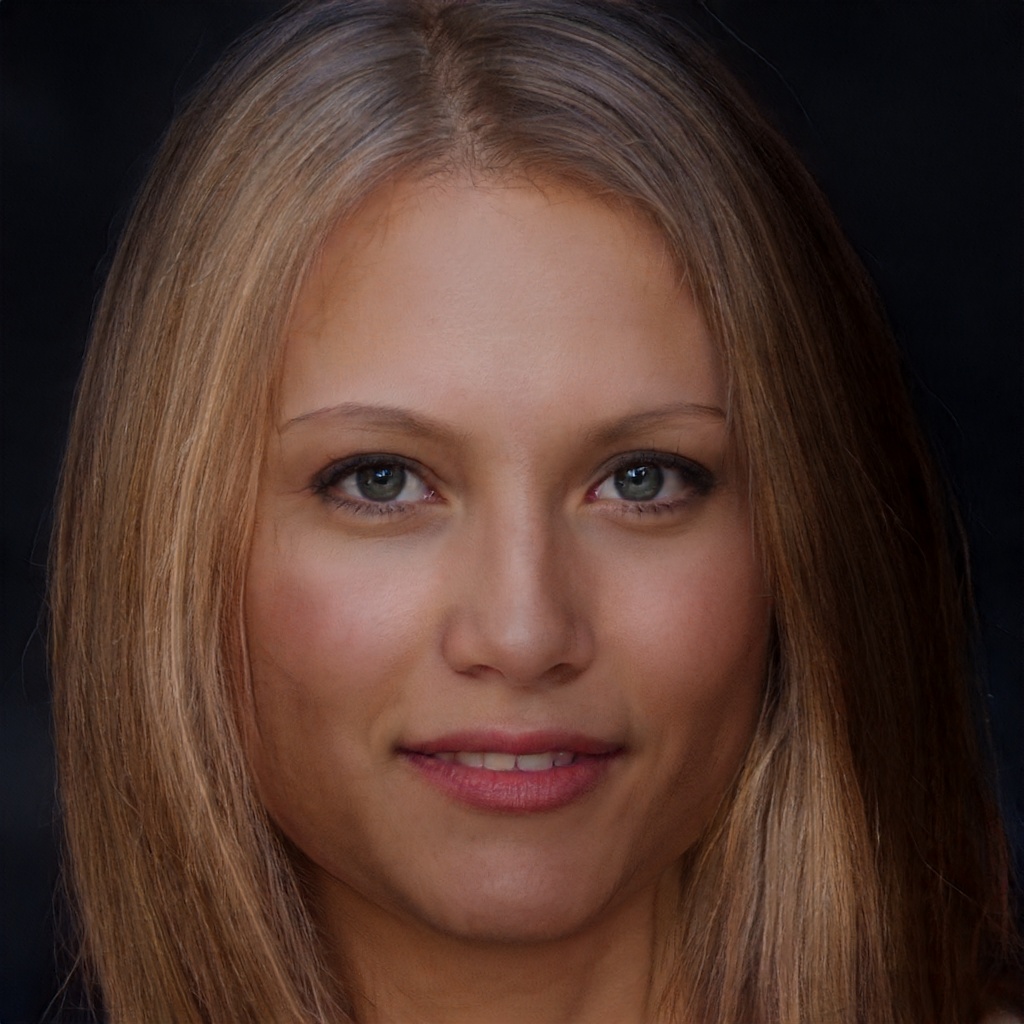}
    \includegraphics[width=1\linewidth]{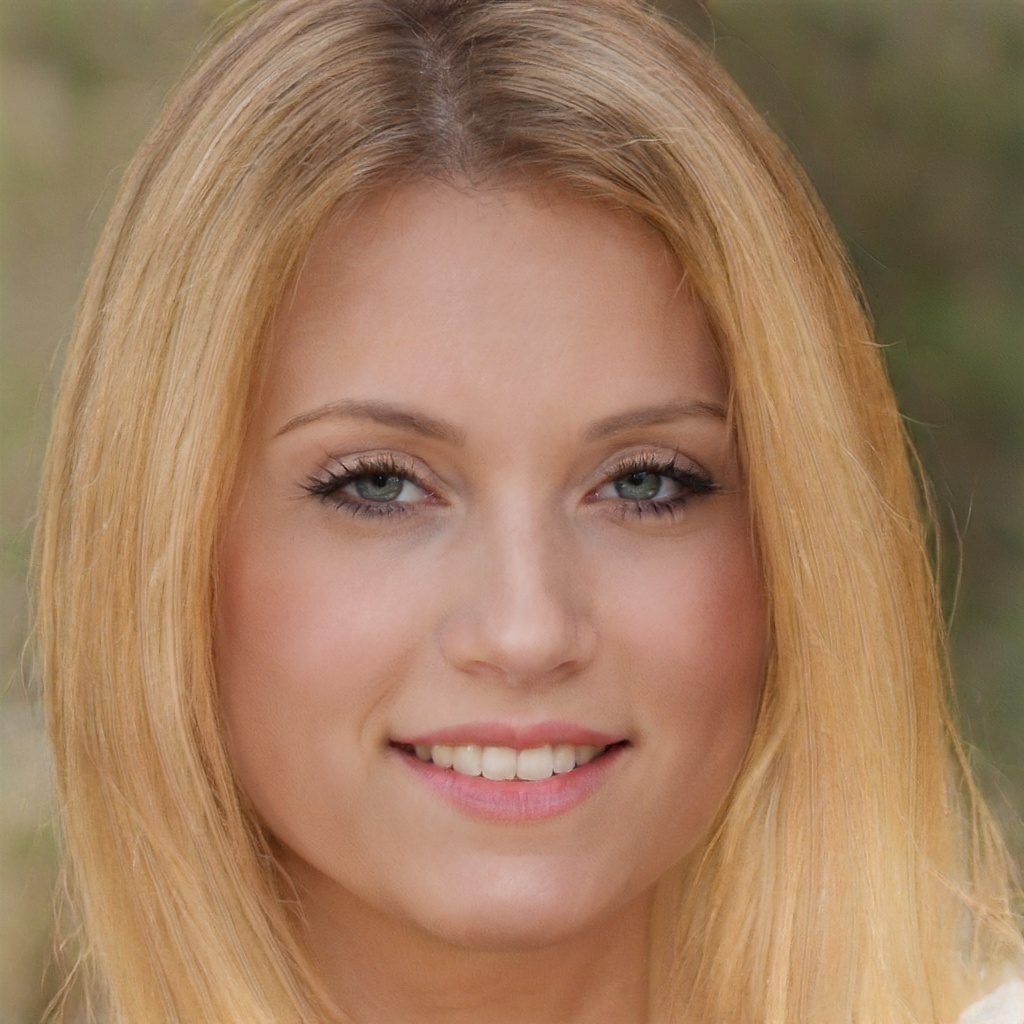}
    }
    \end{minipage}
    \begin{minipage}[b]{0.23\linewidth}
    {
    \begin{center}
        Ours
    \end{center}
    \includegraphics[width=1\linewidth]{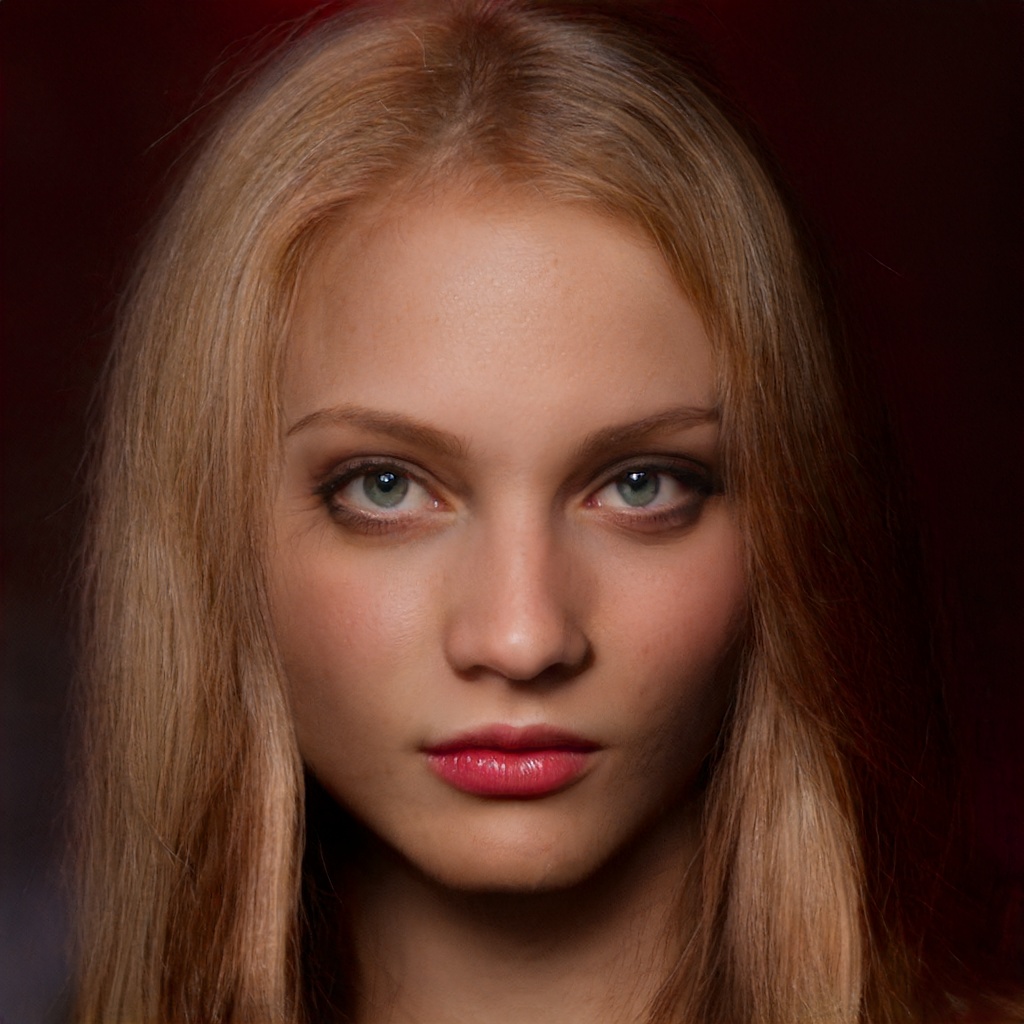}
    \includegraphics[width=1\linewidth]{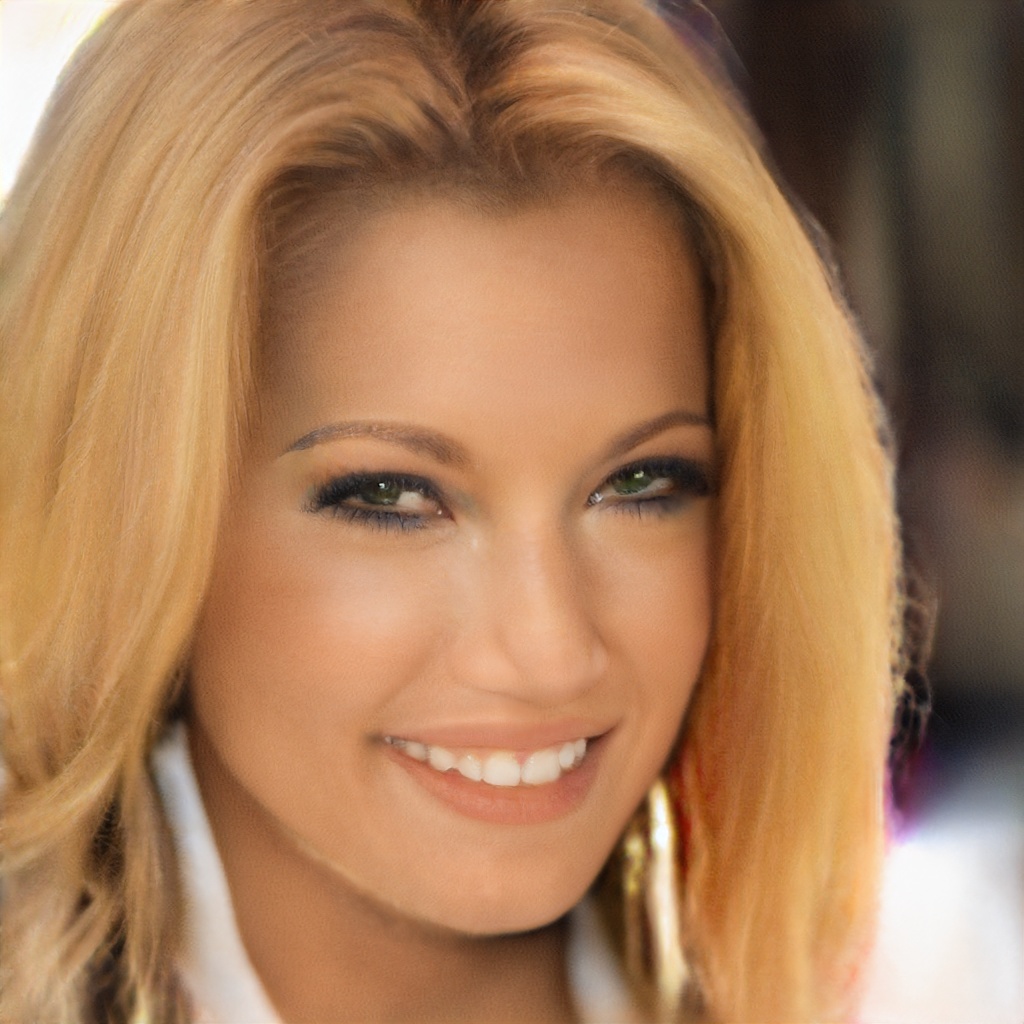}
    }
    \end{minipage}
    \caption{Qualitative results of different regularization strategy. w/o Reg: no regularization term, L1: regularize with L1 norm, Ours: regularize with L2 norm.}
    \label{fig:offset}
\end{figure}

\begin{table}[t]
    \centering
    \caption{Comparison of different regularization strategy. w/o Reg: no regularization term, L1: regularize with L1 norm, Ours: regularize with L2 norm.}
    \label{tab:offset}
    \begin{tabular}{lccc}
    \toprule
        Metric & w/o Reg & $L_1$ & Ours \\
    \midrule
        LPIPS$\downarrow$ & 0.1996 & 0.1694 & \textbf{0.1560}\\ 
        ID Similarity$\uparrow$ & \textbf{0.7251} & 0.1955 & 0.5756\\
    \bottomrule
    \end{tabular}
\end{table}

\paragraph{Effect of Regularized Offset}
To validate the effectiveness of our regularized offset, we conduct an ablation study on face colorization with StyleGAN2. 60 images from CelebA-HQ are used. We first remove the regularization term and then replace the $L_2$ regularization with $L_1$. As shown in Table \ref{tab:offset}, using $L_2$ regularized offset is beneficial for both LPIPS and the ID Similarity compared to $L_1$ regularization. Optimizing without a regularization term (w/o Reg) fails to generate authentic colors (see Figure \ref{fig:offset}) but achieves high ID similarity, which also demonstrates the inherent trade-off between distortion and perception. 
In the meanwhile, our proposed CRI can achieve a ``sweet spot'' in the distortion-perception trade-off by introducing the regularized offset. 

\subsection{Implementation details}
For StyleGAN-XL with ImageNet, we use 1000 iterations for the optimization stage and use 350 iterations for the finetune stage as in \cite{sauer2022stylegan}. 
For StyleGAN2 with FFHQ, we use 500 iterations for the first stage and 20 iterations for the second stage as inversion with a face is simpler. 
For more detailed implementations, please refer to the supplementary materials. 

\section{Conclusion}
We have presented a general framework for generating high-resolution images from degraded images in diverse natural scenes. This is achieved by leveraging the generative power of StyleGAN-XL. Furthermore, we propose a simple yet effective technique, term as Clustering \& Regularize, to ease the inversion difficulty of StyleGAN-XL pre-trained on massive natural images. Clustering is introduced to solve the difficulty caused by the large and complex latent space of StyleGAN-XL. It divides the latent space into sub-spaces and provides the inversion with a better initialization. While Regularize is derived from the perceptual aspect. By introducing an offset term and constraining it with regularization we can bind inversion for better visual quality. 
CRI allows us to achieve good perception and can effectively provide the results with rich image semantics. 
Extensive experiments on image restoration tasks illustrate the effectiveness of CRI. By designing the degradation function $D(\cdot)$, we believe our framework can be applied to other degradations, \textit{e.g.}, noise and blur, which will be left for future work.

\section{Acknowledgments}
This work was supported by the National Key Research and Development Program of China (2021ZD0111000); National Natural Science Foundation of China No. 62222602, 62176092, 62106075, Shanghai Science and Technology Commission No.21511100700, Natural Science Foundation of Shanghai (20ZR1417700).

\bibliography{aaai23}

\end{document}